\documentclass{article}

\usepackage[preprint]{neurips_2020}

\usepackage[utf8]{inputenc} 
\usepackage[T1]{fontenc}    
\usepackage{hyperref}       
\usepackage{url}            
\usepackage{booktabs}       
\usepackage{amsfonts}       
\usepackage{nicefrac}       
\usepackage{microtype}      
\usepackage{graphicx}       
\usepackage{subcaption}
\usepackage{xspace}
\usepackage{amsmath}
\usepackage{algorithm}
\usepackage{algorithmic}

\usepackage{lettrine}

\usepackage{hyperref}

\usepackage{amsthm}
\newtheorem*{example*}{Example}
\usepackage{amsmath}
\usepackage{amssymb}

\usepackage{epsfig}
\usepackage{epstopdf}

\usepackage{graphicx}

\usepackage{xcolor}

\usepackage{comment}

\input{required.tex}
\newcommand{\OTgamma}{{\mathrm{OT}_\gamma}}
\newcommand{\supp}{\mathrm{supp}}
\newcommand{\sinkhorndescent}{{\texttt{Sinkhorn Descent }}}

\newcommand{\sd}{{$\texttt{SD}$ }}

\newcommand{\sod}{{$\texttt{SoD}$}\xspace}
\newcommand{\fw}{{$\texttt{FW}$}\xspace}
\newcommand{\svgd}{{$\texttt{SVGD}$}\xspace}

\newcommand{\FD}{{Fr\'echet derivative }}
\newcommand{\FDs}{{Fr\'echet derivatives }}
\newcommand{\FDcamel}{{Fr\'echet Derivative }}

\usepackage{empheq}

\title{Sinkhorn Barycenter via Functional Gradient Descent}

\author{%
  Zebang Shen$^*$ \quad Zhenfu Wang$^\dagger$ \quad Alejandro Ribeiro$^*$  \quad Hamed Hassani$^*$\\
  $^*$Department of Electrical and Systems Engineering \quad $^\dagger$Department of Mathematics\\
  University of Pennsylvania\\
  \texttt{\{zebang@seas,zwang423@math,aribeiro@seas,hassani@seas\}.upenn.edu}
}

\begin{document}

\maketitle

\begin{abstract}
	In this paper, we consider the problem of computing the barycenter of a set of probability distributions under the Sinkhorn divergence.
	This problem has recently found applications across various domains, including  graphics, learning, and vision, as it provides a meaningful mechanism to aggregate knowledge. 
	Unlike previous approaches which directly operate in the space of probability measures, we recast the Sinkhorn barycenter problem as an instance of unconstrained functional optimization and develop a novel functional gradient descent method named \texttt{Sinkhorn Descent} (\texttt{SD}). 
	We prove that \texttt{SD} converges to a stationary point at a sublinear rate, and under reasonable assumptions, we  further show that it asymptotically finds  a global minimizer of the Sinkhorn barycenter problem.  Moreover, by providing a mean-field analysis, we show that \texttt{SD} preserves the {weak convergence} of empirical measures. 
	Importantly, the computational complexity of \texttt{SD} scales linearly in the dimension $d$ and we demonstrate its scalability by solving a $100$-dimensional Sinkhorn barycenter problem.
\end{abstract}

\section{Introduction} \label{section_introduction}
Computing a nonlinear interpolation between a set of probability measures is a foundational  task
across many disciplines. This problem is typically referred as the barycenter problem and, as it provides a meaningful metric to aggregate knowledge, it has found numerous applications. Examples include distribution clustering \citep{ye2017fast}, Bayesian inference \citep{srivastava2015wasp}, texture mixing \citep{rabin2011wasserstein}, and graphics \citep{solomon2015convolutional}, etc.
The barycenter  problem  can be naturally cast as minimization of the average distance between the target measure (barycenter) and the source measures; and the  choice of the distance metric can significantly impact the quality of the barycenter \citep{feydy2019interpolating}. 
In this regard,  the Optimal Transport (OT) distance (a.k.a. the Wasserstein distance) and its entropy regularized variant (a.k.a. the Sinkhorn divergence) are the most suitable geometrically-faithful metrics, while the latter is more computational friendly.  In this paper, we  provide efficient and provable methods for the Sinkhorn barycenter problem. 

The prior work in this domain has mainly  focused on finding the barycenter by optimizing directly in the space of (discrete) probability measures. We can  divide these previous methods into three broad classes depending on how the support of the barycenter is determined:\\
\indent (i) The first class assumes a fixed and prespecified support set for the barycenter  and only optimizes the corresponding weights \citep{staib2017parallel,dvurechenskii2018decentralize,kroshnin2019complexity}. 
Accordingly, the problem reduces to minimizing a convex objective subject to a simplex constraint. However, fixing the support without any prior knowledge creates undesired bias and affects the quality of the final solution. While increasing the support size (possibly exponentially in the dimension $d$) can help to mitigate the bias, it renders the procedure computationally prohibitive as $d$ grows.\\
\indent (ii) To reduce the bias, the second class considers optimizing the support and the weights through an alternating procedure \citep{cuturi2014fast,claici2018stochastic}.
Since the barycenter objective is not jointly convex with respect to the support and the weights, these methods in general only converge to a stationary point, which can be far from the true minimizers.\\
\indent (iii) Unlike the aforementioned classes, \citet{NIPS2019_9130} recently proposed 
a  conditional gradient method with a growing support set. 
This method enjoys sublinear convergence to the global optimum 
under the premise that a $d$-dimensional nonconvex subproblem can be globally minimized per-iteration.
However,  nonconvex optimization is generally intractable in high dimensional problems (large $d$) and only stationary points can be efficiently reached. 
Hence, the guarantee of  \citep{NIPS2019_9130} has limited applicability as the dimension grows.   

In this paper, we provide a new perspective on the Sinkhorn barycenter problem:  Instead of operating in the space of probability measures, we view the barycenter as the push-forward measure of a given initial measure under an unknown mapping. 
We thus recast the barycenter problem as an unconstrained functional optimization over the space of mappings. Equipped with this perspective, we make the following contributions:
\begin{itemize}
	\item  We develop a novel functional gradient descent method, called \sinkhorndescent (\texttt{SD}), which operates by finding the push-forward mapping in a Reproducing Kernel Hilbert Space that allows the fastest descent, and consequently solves the Sinkhorn barycenter problem iteratively. 
	We then define the Kernelized Sinkhorn Barycenter Discrepancy (KSBD) to characterize the non-asymptotic convergence of \texttt{SD}.
	In particular, we prove that KSBD vanishes under the \texttt{SD} iterates at the rate of $\OM(\frac{1}{t})$, where $t$ is the iteration number.
	\item 
	We prove that \texttt{SD} preserves the weak convergence of empirical measures.
	Concretely, use ${\texttt{SD}}^t(\cdot)$ to denote the output of \texttt{SD} after $t$ iterations and let $\alpha_N$ be an empirical measure of $\alpha$ with $N$ samples.
	We have $\lim_{N\rightarrow\infty}{\texttt{SD}}^t(\alpha_N) = {\texttt{SD}}^t(\alpha)$.
	Such asymptotic analysis allows us to jointly study the behavior of \texttt{SD} under either discrete or continuous initialization.
	\item 
	Under a mild assumption, we prove that KSBD is a valid discrepancy to characterize the optimality of the solution, i.e. the vanishing of  KSBD implies the output measure of \texttt{SD} converges to the global optimal solution set of the Sinkhorn barycenter problem. 
\end{itemize}
Further, we show the efficiency and efficacy of \texttt{SD} by comparing it with prior art on several problems.
We note that the computation complexity of \texttt{SD} depends \emph{linearly} on the dimension $d$. 
We hence validate the scalability of \texttt{SD} by solving a $100$-dimensional barycenter problem, which cannot be handled by previous methods due to their exponential dependence on the problem dimension.
\paragraph{Notations.}
Let $\XM\subseteq\RBB^d$ be a compact ground set, endowed with a symmetric ground metric $c:\XM\times\XM\rightarrow\RBB_+$.
Without loss of generality, we assume $c(x, y) = \infty$ if $x\notin\XM$ or $y \notin \XM$.
We use $\nabla_1 c(\cdot,\cdot):\XM^2\rightarrow \XM$ to denote its gradient w.r.t. its first argument.
Let $\MM_1^+(\XM)$ and $\CM(\XM)$ be the space of probability measures and continuous functions on $\XM$.
We denote the support for a probability measure $\alpha \in \MM_1^+(\XM)$ by $\supp(\alpha)$ and we use $\alpha -a.e.$ to denote "almost everywhere w.r.t. $\alpha$".
For a vector $\aB\in\RBB^d$, we denote its $\ell_2$ norm by $\|\aB\|$.
For a function $f:\XM\rightarrow\RBB$, we denote its $L^\infty$ norm by $\|f\|_\infty \defi \max_{x\in\XM} |f(x)|$ and denote its gradient by $\nabla f(\cdot):\XM\rightarrow\RBB^d$.
For a vector function $f:\XM\rightarrow\RBB^d$, we denote its $(2,\infty)$ norm by $\|f\|_{2, \infty} \defi \max_{x\in\XM} \|f(x)\|$.
For an integer $n$, denote $[n]\defi \{1, \cdots, n\}$.\\
Given an Reproducing Kernel Hilbert Space (RKHS) $\HM$ with a kernel function $k:\XM\times\XM\rightarrow\RBB_+$, we say a vector function $\psi = [[\psi]_1, \cdots, [\psi]_d]\in\HM^d$ if each component $[\psi]_i$ is in $\HM$. The space $\HM$ has a natural inner product structure and an induced norm, and so does  $\HM^d$, i.e. $\langle f, g\rangle_{\HM^d} = \sum_{i=1}^{d} \langle [f]_i, [g]_i\rangle_{\HM}, \forall f, g\in\HM^d$ and the norm $\|f\|_{\HM^d}^2 = {\langle f, f\rangle_{\HM^d}}$. The reproducing property of the RKHS $\HM$ reads that given $f \in \HM^d$, one has $[f]_i(x) = \langle [f]_i, k_x \rangle_{\HM}$ with $k_x(y) = k(x, y)$, which by Cauchy-Schwarz inequality implies that there exists some constant $M_{\HM}>0$ such that
\begin{equation} \label{eqn_RKHS_norm}
	\|f\|_{2, \infty} \leq M_{\HM}\|f\|_{\HM^d}, \forall f\in \HM^d.
	\vspace{-.1cm}
\end{equation}
Additionally, for a functional $F: \HM^d \to \mathbb{R}$, the Fr\'echet derivative of $F$ is defined as follows.
\begin{definition}[Fr\'echet derivative in RKHS] \label{definition_variation_rkhs}
	For a functional $F:\HM^d\rightarrow\RBB$, its Fr\'echet derivative $DF[\psi]$ at $\psi \in\HM^d$ is a function in $\HM^d$ satisfying the following: For any $\xi\in\HM^d$ with $\|\xi\|_{\HM^d}<\infty$,
	$$\lim_{\epsilon\rightarrow0}\frac{F[\psi+\epsilon \xi] - F[\psi]}{\epsilon} = \langle DF[\psi], \xi\rangle_{\HM^d}.$$ 
\end{definition}
Note that the Fr\'echet derivative at $\psi$, i.e. $DF[\psi]$, is a bounded linear operator from $\HM^d$ to $\RBB$.
It can be written in the form $DF[\psi](\xi) = \langle DF[\psi], \xi\rangle_{\HM^d}$ due to the Riesz–Fr\'echet representation theorem.

\subsection{{Related Work on Functional Gradient Descent}}
A related functional gradient descent type method is the Stein Variation Gradient Descent (\svgd) method by \citet{liu2016stein}.
\svgd considers the problem of minimizing the Kullback–Leibler (KL) divergence between a variable distribution and a posterior $p$.
Note that \svgd updates the positions of a set of $N$ particles using the score function of the posterior $p$, i.e. $\nabla \log p$.
Consequently, it requires the access to the target distribution function.
Later, \citet{liu2017stein} prove that \svgd has convergence guarantee in its continuous-time limit (taking infinitesimal step size) using infinite number of particles ($N\rightarrow\infty$). 
In comparison, \texttt{SD} is designed to solve the significantly more complicated Sinkhon barycenter problem and has a stronger convergence guarantee.
More precisely, while \texttt{SD} updates the measure using only a sampling machinery of the target measures (no score functions), it is guaranteed to converge sub-linearly to a stationary point when $\alpha$ is a \emph{discrete} measure using \emph{discrete} time steps.
This is in sharp contrast to the results for \svgd.

In another work, \citet{mroueh2019sobolev} considers minimizing the Maximum Mean Discrepancy (MMD) between a source measure and a variable measure.
They solve this problem by incrementally following a Sobolev critic function and propose the Sobolev Descent (\sod) method.
To show the global convergence of the measure sequence generated by \sod, \citet{mroueh2019sobolev} assumes the \emph{entire} sequence satisfies certain spectral properties, which is in general difficult to verify.
Later, \citet{arbel2019maximum} consider the same MMD minimization problem from a gradient flow perspective.
They propose two assumptions that if either one holds, the MMD gradient flow converges to the global solution.
However, similar to \citep{mroueh2019sobolev}, these assumptions have to be satisfied for the \emph{entire} measure sequence.
We note that the Sinkhorn barycenter is a strict generalization of the above MMD minimization problem and is hence much more challenging: By setting the number of source measures $n=1$ and setting the entropy regularization parameter $\gamma = \infty$, problem \eqref{eqn_sinkhorn_barycenter} degenerates to the special case of MMD.
Further, the MMD between two probability measures has a closed form expression while the Sinkhorn Divergence can only be described via a set of optimization problems. Consequently, the Sinkhorn barycenter is significantly more challenging. To guarantee global convergence, the proposed \sd algorithm only requires one of accumulation points of the measure sequence to be fully supported on $\XM$ with no restriction on the entire sequence.

\section{Sinkhorn Barycenter} \label{section_sinkhorn_barycenter}
We first introduce the entropy-regularized optimal transport distance and its debiased version, a.k.a. the Sinkhorn divergence.
Given two probability measures $\alpha, \beta \in \MM_1^+(\XM)$, 
use $\Pi(\alpha, \beta)$ to denote the set of joint distributions over $\XM^2$ with marginals $\alpha$ and $\beta$. For $\pi\in\Pi$, use $\langle c, \pi\rangle$ to denote the integral $\langle c, \pi\rangle = \int_{\XM^2} c(x, y)\dB\pi(x, y)$ and use $\rm{KL}(\pi||\alpha\otimes\beta)$ to denote the {Kullback-Leibler divergence} between the candidate transport plan $\pi$ and the product measure $\alpha\otimes\beta$.
The entropy-regularized optimal transport distance  $\OTgamma(\alpha, \beta):\MM_1^+(\XM)\times\MM_1^+(\XM)\rightarrow\RBB_+$ is defined as 
\begin{equation}
\OTgamma(\alpha, \beta) = \min_{\pi\in\Pi(\alpha, \beta)} \langle c, \pi\rangle + \gamma \rm{KL}(\pi||\alpha\otimes\beta).
\label{eqn_OTepsilon}
\end{equation}
Here, $\gamma > 0$ is a regularization parameter. Note that $\OTgamma(\alpha, \beta)$ is not a valid metric as there exists $\alpha \in \MM_1^+(\XM)$ such that $\OTgamma(\alpha, \alpha)\neq 0$ when $\gamma\neq 0$.
To remove  this bias, \citet{peyre2019computational} introduced the  \emph{Sinkhorn divergence} $\SBB_\gamma(\alpha, \beta):\MM_1^+(\XM)\times\MM_1^+(\XM)\rightarrow\RBB_+$:
\begin{equation} \label{s-o}
	\SBB_\gamma(\alpha, \beta) \defi \OTgamma(\alpha, \beta) - \frac{1}{2}\OTgamma(\alpha, \alpha) - \frac{1}{2}\OTgamma(\beta, \beta),
\end{equation}
which is a {debiased version} of $\OTgamma(\alpha, \beta)$.
It is further proved that $\SBB_\gamma(\alpha, \beta)$ is nonnegative, bi-convex and metrizes the convergence in law when the ground set $\XM$ is compact and the metric $c$ is Lipschitz.
Now given a set of probability measures $\{\beta_i\}_{i=1}^n$, the Sinkhorn barycenter is the measure $\alpha\in \MM_1^+(\XM)$ that minimizes the average of Sinkhorn divergences
\begin{equation}
	\min_{\alpha\in \MM_1^+(\XM)}  \Big( \SM_{\gamma}(\alpha)\defi\frac{1}{n}\sum_{i=1}^{n}\SBB_\gamma(\alpha, \beta_i) \Big).
	\label{eqn_sinkhorn_barycenter}
\end{equation}
We will next focus on the properties of $\OTgamma$ since $\SM_{\gamma}(\alpha)$ is the linear combination of these terms.

\paragraph{The Dual Formulation of $\OTgamma$.} As a convex program, the entropy-regularized optimal transport problem $\OTgamma$ \eqref{eqn_OTepsilon} has a equivalent dual formulation, which is given as follows:
\begin{align} \label{eqn_OTepsilon_dual} 
\OTgamma(\alpha, \beta) = \max_{f, g\in\CM(\XM)} \langle f, \alpha\rangle 
+ \langle g, \beta\rangle  - \gamma\langle \exp((f\oplus g - c)/\gamma) - 1, \alpha\otimes\beta\rangle,
\end{align}
where we denote $[f\oplus g](x, y) = f(x) + g(y)$.
The maximizers $f_{\alpha, \beta}$ and  $g_{\alpha, \beta}$ of \eqref{eqn_OTepsilon_dual} are called the \emph{Sinkhorn potentials} of $\OTgamma(\alpha, \beta)$.
Define the Sinkhorn mapping $\AM:\CM(\XM)\times\MM_1^+(\XM) \rightarrow \CM(\XM)$ by
\begin{equation} \label{eqn_Sinkhorn_mapping}
\AM(f, \alpha)(y) = -\gamma\log\int_\XM\exp\big(({f(x) - c(x, y)})/{\gamma}\big)\dB\alpha(x).
\end{equation}
The following lemma states the optimality condition for the Sinkhorn potentials $f_{\alpha, \beta}$ and $g_{\alpha, \beta}$.
\begin{lemma}[Optimality \cite{peyre2019computational}] \label{lemma_optimality_sinkhorn_potential}
	The pair $(f, g)$ are the Sinkhorn potentials of the entropy-regularized optimal transport problem \eqref{eqn_OTepsilon_dual} if they satisfy 
	\begin{equation}
	f = \AM(g, \beta), \alpha-a.e. \quad \textrm{and}\quad g = \AM(f, \alpha), \beta-a.e. \label{eqn_optimality_sinkhorn_potential_xy}.
	\end{equation}
\end{lemma}

The Sinkhorn potential is the cornerstone of the entropy regularized OT problem. In the discrete case, it can be computed by a standard method  in \cite{genevay2016stochastic}. In particular, when $\alpha$ is discrete, $f$ can be simply represented by a finite dimensional vector since only its values on $\supp(\alpha)$ matter. We describe such method in Appendix \ref{section_computation_of_sinkhorn_potential} for completeness.
In the following, we treat the computation of Sinkhorn potentials as a blackbox, and refer to it as $\mathcal{SP}_{\gamma}(\alpha,\beta)$.


\section{Methodology}
We present the \sinkhorndescent (\texttt{SD}) algorithm for the Sinkhorn barycenter problem \eqref{eqn_sinkhorn_barycenter} in two steps:
We first reformulate \eqref{eqn_sinkhorn_barycenter} as an unconstrained functional minimization problem and then derive the descent direction as the negative functional gradient over a RKHS $\HM^d$.
Operating in RKHS allows us to measure the quality of the iterates  using a so-called kernelized discrepancy which we introduce in Definition \ref{definition_KSBD}.  This quantity  will be crucial for our convergence analysis.
The restriction of a functional optimization problem to RKHS is common in the literature as discussed in Remark \ref{remark_RKHS}. 
\paragraph{Alternative Formulation.}
Instead of directly solving the Sinkhorn barycenter problem in the probability space $\MM_1^+(\XM)$, we reformulate it as a functional minimization over all mappings on $\XM$: 
\begin{equation}
\min_{\PM}  \Big( \SM_{\gamma}(\PM_\sharp\alpha_0) \defi\frac{1}{n}\sum_{i=1}^{n}\SBB_\gamma(\PM_\sharp\alpha_0, \beta_i) \Big),
\label{eqn_sinkhorn_barycenter_transform}
\end{equation}
where $\alpha_0\in\MM_1^+(\XM)$ is some given initial measure, and $\PM_\sharp\alpha$ is the push-forward measure of $\alpha\in\MM_1^+(\XM)$ under the mapping $\PM:\XM\rightarrow\XM$.
When $\alpha_0$ is sufficiently regular, e.g. absolutely continuous, for any $\alpha\in\MM_1^+(\XM)$ there always exists a mapping $\PM$ such that $\alpha = \PM_\sharp\alpha_0$ (see Theorem 1.33 of \citep{ambrosio2013user}).
Consequently, problems \eqref{eqn_sinkhorn_barycenter_transform} and \eqref{eqn_sinkhorn_barycenter} are equivalent with appropriate initialization.
\paragraph{Algorithm Derivation.}
\begin{algorithm}[tb]
	\caption{\sinkhorndescent (\texttt{SD})}
	\label{alg:Sinkhonr_Descent_Finite_Particles}
	\begin{algorithmic}
		\STATE {\bfseries Input:} measures $\{\beta_i\}_{i=1}^n$, a discrete initial measure $\alpha^{0}$, a step size $\eta$, and number of iterations $S$;
		\STATE {\bfseries Output:} A measure $\alpha^{S}$ that approximates the Sinkhorn barycenter of $\{\beta_i\}_{i=1}^n$;
		\FOR{$t = 0$ {\bfseries to} $S-1$}
		\STATE $\alpha^{t+1} := \TM[\alpha^{t}]_\sharp\alpha^{t}$, with $\TM[\alpha^{t}]$ defined in \eqref{eqn_pushforward_mapping};
		\ENDFOR
	\end{algorithmic}
\end{algorithm}
For a probability measure $\alpha$, define the functional $\SM_\alpha:\HM^d\rightarrow\RBB$
\begin{equation} \label{eqn_functional_per_iteration}
\SM_\alpha[\psi] = \SM_{\gamma}\big({(\IM + \psi)}_\sharp\alpha\big), \psi\in\HM^d.
\end{equation}
Here $\IM$ is the identity mapping and $\SM_{\gamma}$ is defined in \eqref{eqn_sinkhorn_barycenter}.
Let $\alpha^t$ be the estimation of the Sinkhorn barycenter in the $t^{th}$ iteration.
\sinkhorndescent (\texttt{SD}) iteratively updates the measure $\alpha^{t + 1}$ as 
\begin{equation}
\alpha^{t+1} = {\TM[\alpha^{t}]}_\sharp\alpha^{t}, \label{eqn_sequence_of_measures}
\end{equation}  
via the push-forward mapping (with $\eta>0$ being a step-size)
\begin{equation} \label{eqn_pushforward_mapping}
\TM[\alpha^{t}](x) = x - \eta\cdot D\SM_{\alpha^{t}}[0](x).
\end{equation}
Recall that $D\SM_\alpha[0]$ is the Fr\'echet derivative of $\SM_\alpha$ at $\psi = 0$ (see Definition \ref{definition_variation_rkhs}).
Note that $(\IM+\psi)_\sharp \alpha = \alpha$ when $\psi = 0$.
Our choice of the negative Fr\'echet derivative in $\TM[\alpha^{t}]$ allows the objective $\SM_\gamma(\alpha)$ to have the fastest descent at the current measure $\alpha = \alpha^t$.
We our line the details of \texttt{SD} in Algorithm~\ref{alg:Sinkhonr_Descent_Finite_Particles}.
Consequently, a solution of \eqref{eqn_sinkhorn_barycenter_transform} will be found by finite-step compositions and then formally passing to the limit 
	$\PM = \lim_{t\rightarrow\infty} \left(\PM^t \defi \TM[\alpha^t]\circ\cdots\circ \TM[\alpha^0] \right)$.
\begin{remark} \label{remark_RKHS}
	We restrict  $\psi$ in \eqref{eqn_functional_per_iteration} to the space $\HM^d$ to avoid the inherent difficulty when the perturbation of Sinkhorn potentials introduced by the mapping $(\IM+\psi)$ can no longer be properly bounded (for $\psi \in \HM^d$, we always have the upper bound \eqref{eqn_RKHS_norm} which is necessary in our convergence analysis).
	This restriction  will potentially introduce error to the minimization of \eqref{eqn_sinkhorn_barycenter_transform}.
	However, this restriction is a common practice for general functional optimization problems: 
	Both \svgd \citep{liu2016stein} and \sod \citep{mroueh2019sobolev} explicitly make such RKHS restriction on their transport mappings.
	\citep{arbel2019maximum} constructs the transport mapping using the witness function of the Maximum Mean Discrepancy (MMD) which also lies in an RKHS.
\end{remark}
In what follows, we first derive a formula for the Fr\'echet derivative $D\SM_{\alpha^{t}}[0]$ (see \eqref{eqn_gradient_sinkhorn_barycenter}) and then explain how it is efficiently computed.
The proof of the next proposition requires additional continuity study of the Sinkhorn potentials and is deferred to Appendix \ref{proof_proposition_variation_I}.
\begin{proposition} \label{proposition_variation_I}
	Recall the Fr\'echet derivative in Definition \ref{definition_variation_rkhs}.
	Given $\alpha,\beta\in\MM_1^+(\XM)$, for $\psi\in\HM^d$ denote $F_1[\psi] = \OTgamma\big({(\IM+\psi)}_\sharp\alpha, \beta\big)$ and $F_2[\psi] = \OTgamma\big({(\IM+\psi)}_\sharp\alpha, {(\IM+\psi)}_\sharp\alpha\big)$.
	Under Assumptions \ref{ass_c} and \ref{ass_k} (described below), we can compute
	\begin{align}
		DF_1[0](y) = \int_\XM \nabla f_{\alpha,\beta}(x) k(x, y) \dB \alpha(x), \quad
		DF_2[0](y) = 2\int_\XM \nabla f_{\alpha,\alpha}(x) k(x, y) \dB \alpha(x), \label{eqn_proposition_variation_I} 
	\end{align}
	where $\nabla f_{\alpha,\beta}$ and $\nabla f_{\alpha,\alpha}$ are the gradients of the Sinkhorn potentials of $\OTgamma(\alpha, \beta)$ and $\OTgamma(\alpha, \alpha)$ respectively, and $k$ is the kernel function of the RKHS $\HM$.
\end{proposition}
Consequently the \FD of the Sinkhorn Barycenter problem \eqref{eqn_functional_per_iteration} can be computed by
\begin{align}
D\SM_{\alpha}[0](y)=\int_\XM \frac{1}{n}[\sum_{i=1}^{n}\nabla f_{\alpha,\beta_i}(x) -
\nabla f_{\alpha, \alpha}(x)] k(x, y) \dB \alpha(x). \label{eqn_gradient_sinkhorn_barycenter}
\end{align}
This quantity can be computed efficiently when $\alpha$ is discrete: Consider an individual term $\nabla f_{\alpha,\beta}$.
Define 
$h(x, y) \defi \exp\left(\frac{1}{\gamma}(f_{\alpha,\beta}(x) + \AM[f_{\alpha,\beta}, \alpha](y) - c(x, y))\right)$. Lemma \ref{lemma_optimality_sinkhorn_potential} implies $$\int h(x, y)\dB\beta(y) = 1.$$
Taking derivative with respect to $x$ on both sides and rearranging terms, we have
\begin{align}
\nabla f_{\alpha,\beta}(x) = \frac{\int_\XM h(x, y) \nabla_x c(x, y) \dB\beta(y)}{\int h(x, y)\dB\beta(y)} = \int_\XM h(x, y)\nabla_x c(x, y) \dB\beta(y) \label{eqn_sinkhorn_potential_gradient_x},
\end{align}
which itself is an expectation.
Note that to evaluate \eqref{eqn_gradient_sinkhorn_barycenter}, we only need $\nabla f_{\alpha,\beta}(x)$ on $\supp(\alpha)$.
Using $\mathcal{SP}_{\gamma}(\alpha,\beta)$ (see the end of Section \ref{section_sinkhorn_barycenter}), the function value of $f_{\alpha,\beta}$ on $\supp(\alpha)$ can be efficiently computed.
Together with the expression in \eqref{eqn_sinkhorn_potential_gradient_x}, the gradients $\nabla f_{\alpha,\beta}(x)$ at $x\in \supp(\alpha)$ can also be obtained by a simple Monte-Carlo integration with respect to $\beta$.
\section{Analysis}
In this section, we analyze the finite time convergence and the mean field limit of $\texttt{SD}$ under the following assumptions on the ground cost function $c$ and the kernel function $k$ of the RKHS $\HM^d$.
\begin{assumption}\label{ass_c}
	The ground cost function $c(x, y)$ is bounded, i.e. $\forall x,y \in \XM, c(x, y) \leq M_c$; $G_c$-Lipschitz continuous, i.e. $\forall x, x', y \in \XM, |c(x, y) - c(x', y)|\leq G_c\|x - x'\|$; and $L_c$-Lipschitz smooth, i.e. $\forall x, x', y \in \XM, \|\nabla_1 c(x, y) - \nabla_1 c(x', y)\|\leq L_c\|x - x'\|$.
\end{assumption}
\begin{assumption}\label{ass_k}
	The kernel function $k(x, y)$ is bounded, i.e. $\forall x,y \in \XM, k(x, y) \leq D_k$; $G_k$-Lipschitz continuous, i.e. $\forall x, x', y \in \XM, |k(x, y) - k(x', y)|\leq G_c\|x - x'\|$.
\end{assumption}
\subsection{Finite Time Convergence Analysis}
In this section, we prove that \sinkhorndescent converges to a stationary point of problem \eqref{eqn_sinkhorn_barycenter} at the rate of $\OM(\frac{1}{t})$, where $t$ is the number of iterations.
We first introduce a discrepancy quantity.
\begin{definition}\label{definition_KSBD}
	Recall the definition of the functional $\SM_{\alpha}$ in \eqref{eqn_functional_per_iteration} and the definition of \FD in Definition \ref{definition_variation_rkhs}.
	Given a probability measure $\alpha\in\MM_1^+(\XM)$, the { Kernelized Sinkhorn Barycenter Discrepancy} (KSBD) for the Sinkhorn barycenter problem is defined as
	\begin{align}\label{ExtraWang1}
	\SB(\alpha, \{\beta_i\}_{i=1}^n) \defi \|D\SM_{\alpha}[0]\|^2_{\HM^d}.
	\end{align}
\end{definition}
Note that in each round $t$, $\SB(\alpha^t, \{\beta_i\}_{i=1}^n)$ metrizes the stationarity of \texttt{SD}, which can be used to quantify the per-iteration improvement.
\begin{lemma}[Sufficient Descent] \label{lemma_sufficient_descent}
	Recall the definition of the Sinkhorn Barycenter problem in \eqref{eqn_sinkhorn_barycenter} and the sequence of measures $\{\alpha^t\}_{t\geq 0}$ 
	in \eqref{eqn_sequence_of_measures} generated by \texttt{SD} (Algorithm \ref{alg:Sinkhonr_Descent_Finite_Particles}). 
	Under Assumption \ref{ass_c}, if we have $\eta\leq \min\{{1}/({8L_fM_\HM^2}), {1}/({8\sqrt{d}L_TM_\HM^2})\}$, the Sinkhorn objective always decreases, 
	\begin{equation}
		\SM_{\gamma}(\alpha_{t+1}) - \SM_{\gamma}(\alpha_{t}) \leq - {\eta}/{2}\cdot\SB(\alpha^t, \{\beta_i\}_{i=1}^n).
	\end{equation}
	See $M_\HM$ in \eqref{eqn_RKHS_norm}, $L_f \defi {4G_c^2}/{\gamma}+L_c$ and $L_T\defi2 G_c^2\exp(3M_c/\gamma)/\gamma$\footnote{We acknowledge the factor $\exp(1/\gamma)$ is non-ideal, but such quantity constantly appears in the literature related to the Sinkhorn divergence, e.g. Theorem 5 in \citep{NIPS2019_9130} and Theorem 3 in \citep{genevay2019sample}. It would be an interesting future work to remove this factor.}.
\end{lemma}
The proof of the lemma in given Appendix \ref{proof_lemma_sufficient_descent}.
Based on this result, we can  derive the following convergence result demonstrating that \texttt{SD} converges to a stationary point in a sublinear rate.
\begin{theorem}[Convergence] \label{theorem_convergence}
	Suppose \texttt{SD} is initialized with $\alpha^0\in\MM_1^+(\XM)$ and outputs $\alpha^t\in\MM_1^+(\XM)$ after $t$ iterations.
	Under Assumption \ref{ass_c}, we have
	\begin{equation}
		\min_{t} \SB(\alpha^t, \{\beta_i\}_{i=1}^n) \leq {2\SM_{\gamma}(\alpha^0)}/{(\eta t)},
	\end{equation}
	where $0<\eta\leq \min\{{1}/({8L_fM_\HM^2}), {1}/({8\sqrt{d}L_TM_\HM^2})\}$ is the step size.
\end{theorem}
With a slight change to \texttt{SD}, we can conclude its last term convergence as elaborated in Appendix \ref{appendix_last_term_convergence}.
\subsection{Mean Field Limit Analysis} \label{section_mean_field_limit}
While \sinkhorndescent accepts both discrete and continuous measures as initialization, in practice, we start from a discrete initial measure $\alpha_N^0$ with $|\supp(\alpha^0_N)| = N$.
If $\alpha_N^0$ is an empirical measure sampled from an underlying measure $\alpha^0_{\infty}$, we have the weak convergence at time $t=0$, i.e. $\alpha_N^0\rightharpoonup\alpha^0_{\infty}$ as $N\rightarrow \infty$.
The mean field limit analysis demonstrates that \sinkhorndescent preserves such weak convergence for any finite time $t$:
\begin{equation*}
	\alpha_N^0\rightharpoonup\alpha^0_{\infty} \Rightarrow \alpha_N^{t} = {\texttt{SD}}^t(\alpha_N^0)\rightharpoonup\alpha_\infty^{t} = {\texttt{SD}}^t(\alpha^0_{\infty}),
\end{equation*}
where we use ${\texttt{SD}}^t$ to denote the output of \texttt{SD} after $t$ steps and use $\rightharpoonup$ to denote the weak convergence.
\begin{lemma}
	Recall the push-forward mapping $\TM[\alpha](x)$ in \texttt{SD} from \eqref{eqn_pushforward_mapping} and 
	recall $L_f$ in Lemma \ref{lemma_sufficient_descent}.
	Under Assumptions \ref{ass_c} and \ref{ass_k}, for two probability measures $\alpha$ and $\alpha'$, we have
	\begin{align}
	d_{bl}(\TM[\alpha]_\sharp \alpha, \TM[\alpha']_\sharp \alpha') \leq (1+\eta C)d_{bl}(\alpha, \alpha'),
	\end{align}
	where $C = G_cG_k+\max\{d L_f D_k + dG_c G_k, {D_k L_{bl}}\}$ and $L_{bl}\defi 8G_c^2\exp(6M_c/\gamma)$.
	\label{lemma_large_N}
\end{lemma}
The proof is presented in Appendix \ref{proof_lemma_large_N}.
This is a discrete version of Dobrushin's estimate (section 1.4. in \citep{golse2016dynamics}).
As a result, we directly have the following large N characterization of ${\texttt{SD}}^t(\alpha^0_N)$.
\begin{theorem}[Mean Field Limit]
	Let $\alpha^0_N$ be an empirical initial measure with $|\supp(\alpha^0_N)| = N$ and let $\alpha^0_\infty$ be the underlying measure such that $\alpha_N^0 \rightharpoonup \alpha^0_{\infty}$.
	Use ${\texttt{SD}}^t(\alpha_N^0)$ and ${\texttt{SD}}^t(\alpha^0_\infty)$ to denote the outputs of \texttt{SD} after $t$ iterations, under the initializations $\alpha_N^0$ and $\alpha^0_\infty$ respectively.
	Under Assumptions \ref{ass_c} and \ref{ass_k}, for any finite time $t$, we have
	\begin{equation*}
		d_{bl}({\texttt{SD}}^t(\alpha_N^0), {\texttt{SD}}^t(\alpha^0_{\infty})) \leq (1+\eta C)^t d_{bl}(\alpha_N^0, \alpha_\infty^0),
	\end{equation*}
	and hence as $N\rightarrow\infty$ we have
	\begin{equation}
		\alpha_N^{t} = {\texttt{SD}}^t(\alpha_N^0) \rightharpoonup \alpha_\infty^{t} = {\texttt{SD}}^t(\alpha^0_{\infty}).
	\end{equation}
\end{theorem}
\subsection{KSBD as Discrepancy Measure}
In this section, we show that, under additional assumptions, KSBD is a valid discrepancy measure, i.e. $\SM_{\gamma}(\alpha) = 0$ implies that $\alpha$ is a global optimal solution to the Sinkhorn barycenter problem \eqref{eqn_sinkhorn_barycenter}. The proof is provided in Appendix \ref{appendix_global_optimality}.
First, we introduce the following positivity condition.
\begin{definition}
	A kernel $k(x, x')$ is said to be integrally strictly positive definite (ISPD) w.r.t. a measure $\alpha\in\MM_1^+(\XM)$, if $\forall \xi:\XM\rightarrow\RBB^d$ with $0<\int_{\XM}\|\xi(x)\|^2 \dB \alpha(x)<\infty$, it holds that 
	\begin{equation}
		\int_{\XM^2} \xi(x) k(x, x') \xi(x') \dB \alpha(x)\dB \alpha(x') >0. \label{eqn_kernel_positivity}
	\end{equation}
\end{definition}
\begin{theorem} \label{thm_optimality} Recall the \FD of the Sinkhorn Barycenter problem in \eqref{eqn_gradient_sinkhorn_barycenter} and KSBD in \eqref{ExtraWang1}.
Denote $\xi(x)\defi \frac{1}{n}\sum_{i=1}^{n} \big(\nabla f_{\alpha, \beta_i}(x) - \nabla f_{\alpha, \alpha}(x)\big)$. We have $\int_{\XM}\|\xi(x)\|^2 \dB \alpha(x)<\infty$.\\
(i) If  the kernel function $k(x, x')$ is ISPD w.r.t. $\alpha\in\MM_1^+(\XM)$ and $\alpha$ is fully supported on $\XM$, then the vanishing of KSBD, i.e. $\SB(\alpha, \{\beta_i\}_{i=1}^n) = 0$, implies that  $\alpha$ globally minimizes problem \eqref{eqn_sinkhorn_barycenter}.\\
(ii) Use ${\alpha}^t$ to denote the output of \sd after $t$ iterations.
If further  one of the accumulation points of the sequence $\{\alpha^t\}$ is fully supported on $\XM$, then $\lim_{t\rightarrow\infty} \SM_{\gamma}(\alpha^t) = \SM_{\gamma}(\alpha^*)$.
\end{theorem}
We show in Appendix \ref{appendix_fully_supported}, under an absolutely continuous (a.c.) and fully supported (f.s.) initialization, $\alpha^t$ remains a.c. and f.s. for any finite $t$. 
This leads to our assumption in (ii): One of the accumulation points of $\{\alpha^t\}$ is f.s..
However, to rigorously analyze the support of $\alpha^t$ in the asymptotic case ($t\rightarrow \infty$) requires a separate proof.
Establishing the global convergence of the functional gradient descent is known to be difficult in the literature, even for some much easier settings compared to our problem \eqref{eqn_sinkhorn_barycenter}.
For instance, \citep{mroueh2019sobolev,arbel2019maximum} prove the global convergence of their MMD descent algorithms. Both works require additional assumptions on the \emph{entire} measure sequence $\{\alpha^t\}$ as detailed in Appendix \ref{appendix_previous_work_assumption}. See also the convergence analysis of SVGD in \citep{lu2019scaling} under very strong assumptions of the score functions. 
\section{Experiments}


We conduct experimental studies to show the efficiency and efficacy of \sinkhorndescent by comparing with the recently proposed functional Frank-Wolfe method (\fw) from \citep{NIPS2019_9130}\footnote{ \citep{claici2018stochastic} is not included as it only applies to the Wasserstein barycenter problem ($\gamma = 0$).}.
Note that in round $t$, \fw requires to globally minimize the nonconvex function 
$Q(x) \defi \sum_{i=1}^{n} f_{\alpha^t, \beta_i}(x) - f_{\alpha^t, \alpha^t}(x)$
in order to choose the next Dirac measure to be added to the support.
Here, $f_{\alpha^t, \beta_i}$ and $f_{\alpha^t, \alpha^t}$ are the Sinkhorn potentials.
Such operation is implemented by an exhaustive grid search so that \fw returns a reasonably accurate solution.
Consequently, \fw is computationally expensive even for low dimensional problems and we only compare \texttt{SD} with \fw in the first two image experiments, where $d = 2$. (the grid size used in \fw grows exponentially with $d$.)\\
Importantly, the size of the support $N$ affects the computational efficiency as well as the solution quality of both methods.
A large support size usually means higher computational complexity but allows a more accurate approximation of the barycenter.
However, since \texttt{SD} and \fw have different support size patterns, it is hard to compare them directly:
The support size of \texttt{SD} is fixed after its initialization while \fw starts from an initial small-size support and gradually increases it during the optimization procedure.
We hence fix the support size of the output measure from \fw and vary the support size of \texttt{SD} for a more comprehensive comparison.
\begin{figure*}[t]
	\centering
	\begin{tabular}{c c c}
		\includegraphics[width=.3\columnwidth]{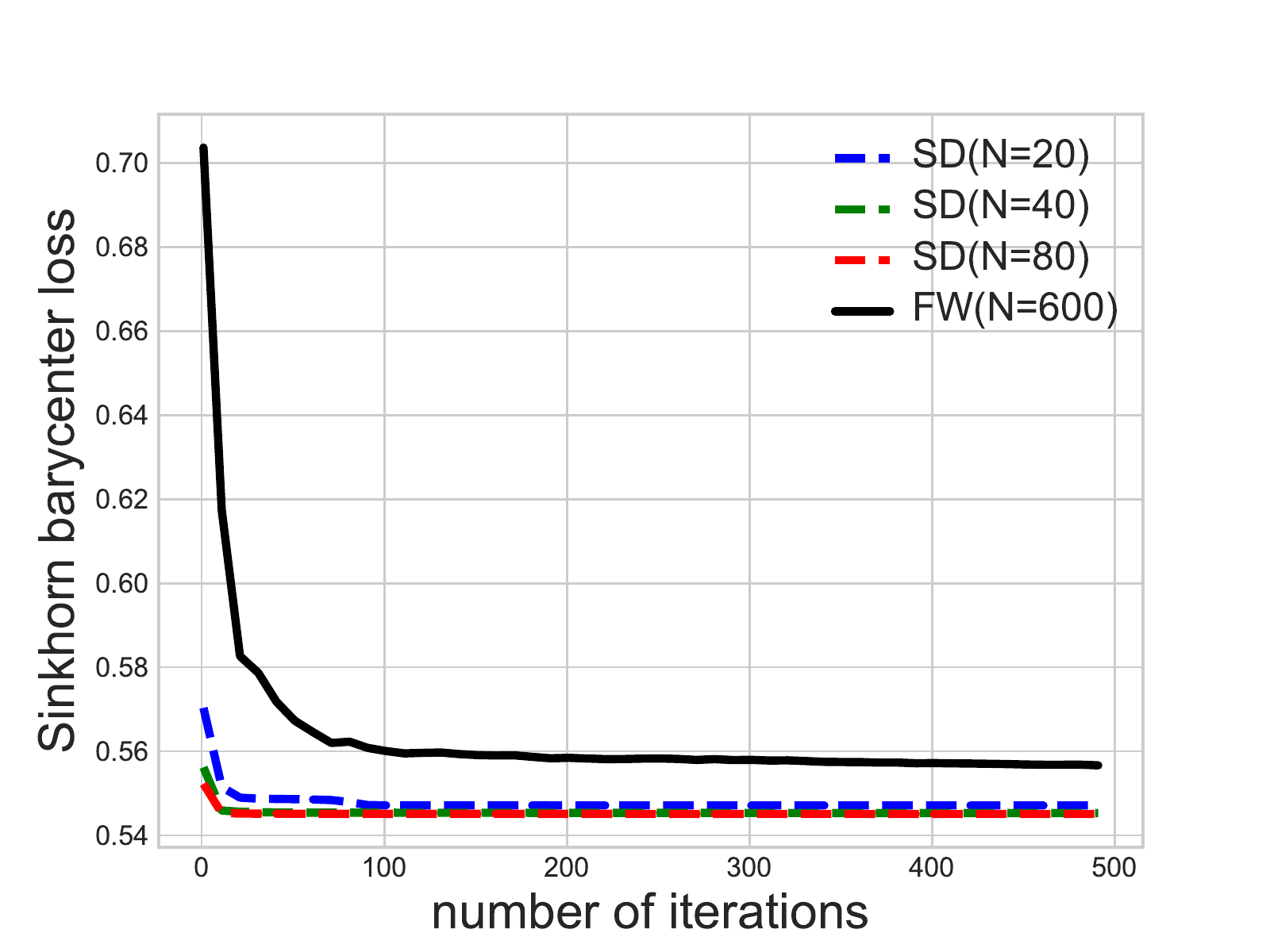} & 	\includegraphics[width=.3\columnwidth]{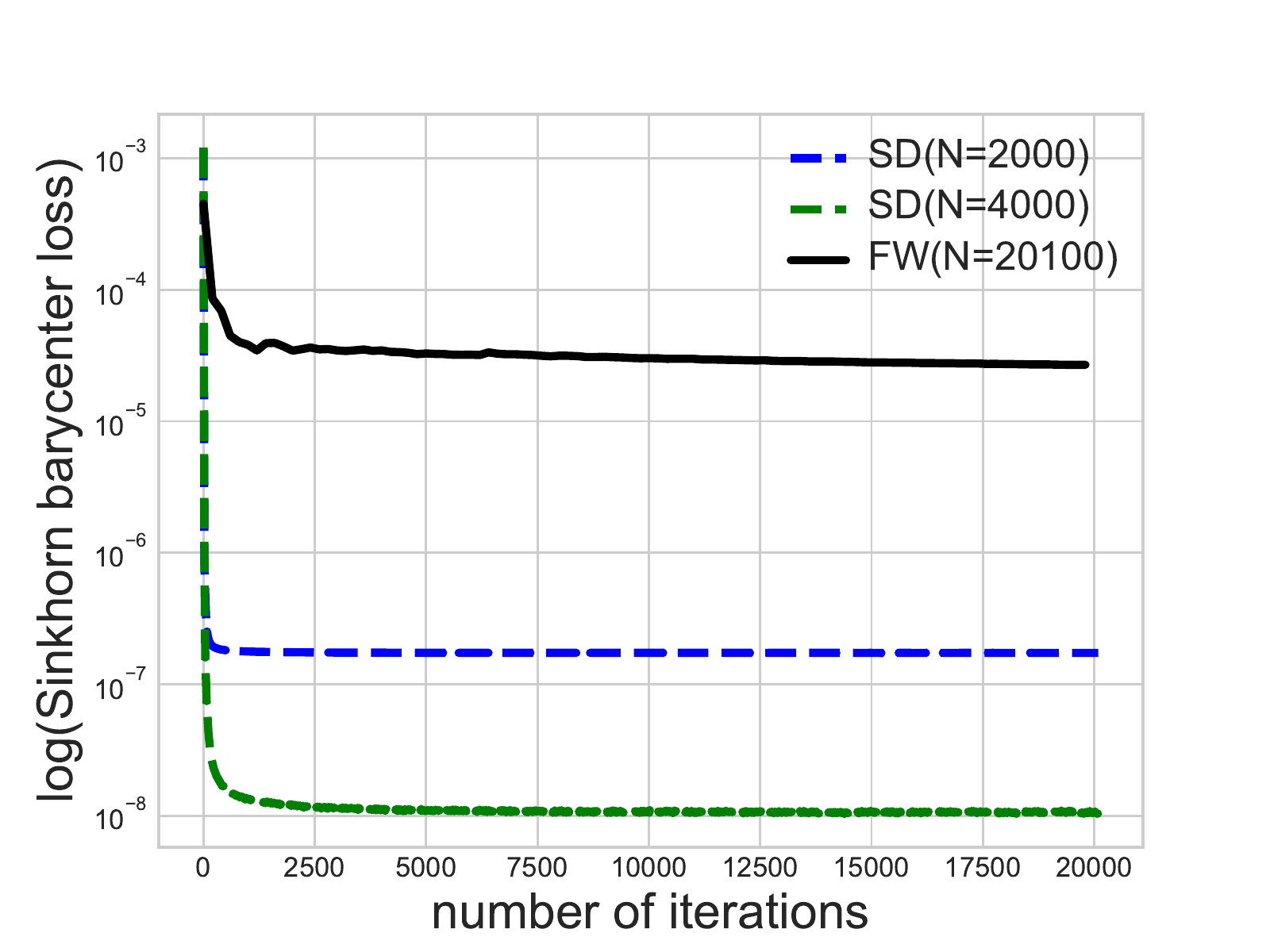} & 	\includegraphics[width=.3\columnwidth]{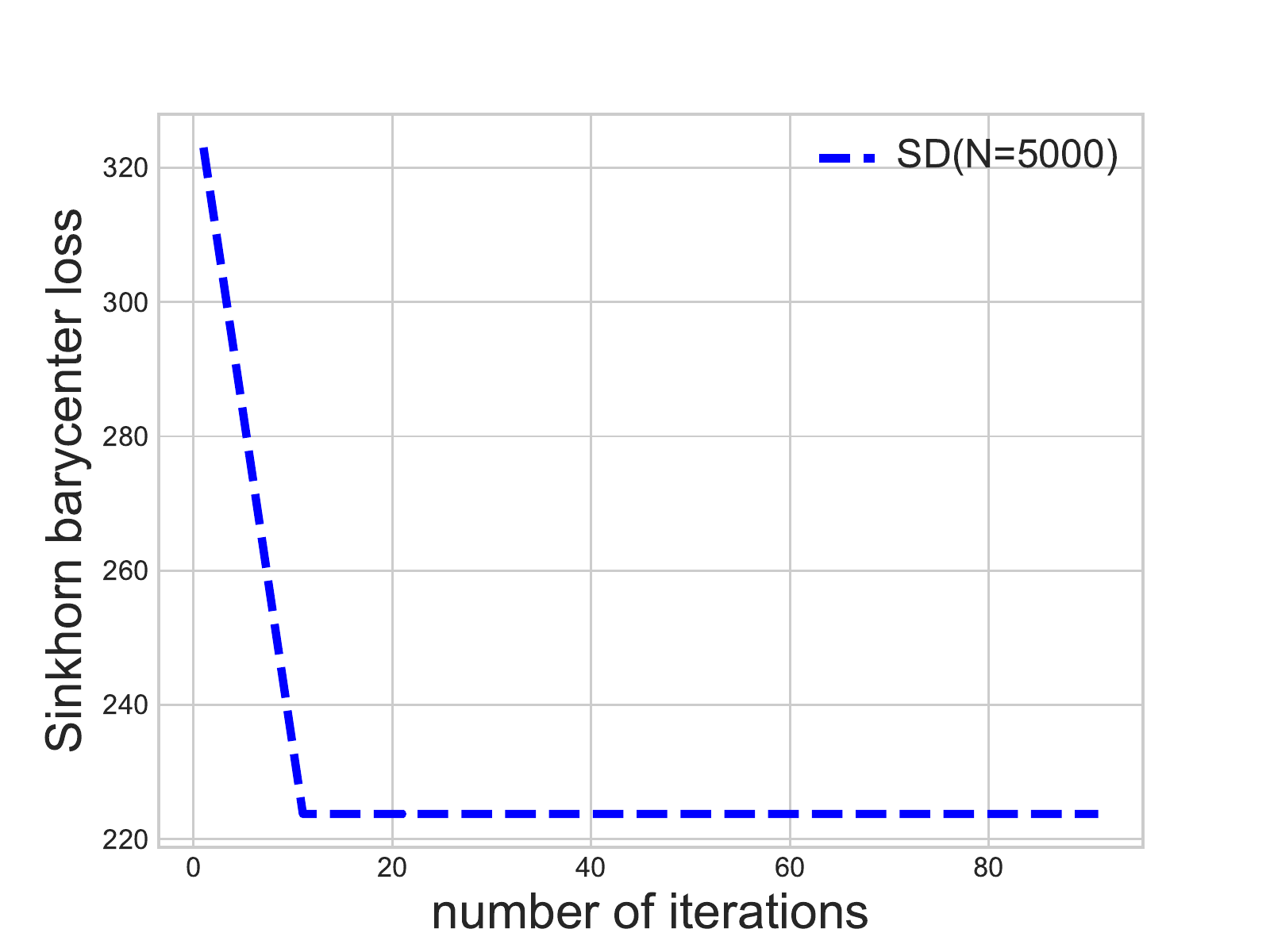}\\
		(a) Concentric Ellipses & (b) Distribution Sketching & (c) Gaussians
	\end{tabular}
	\caption{$N$ is the support size. \fw is not included in (c) as it is impractical in high-dimensional problems (here, the dimension is $100$)}
	\label{fig}
\end{figure*}
\begin{figure*}[t]
	\centering
	\begin{tabular}{c c}
		\includegraphics[height=.08\columnwidth, width=.08\columnwidth]{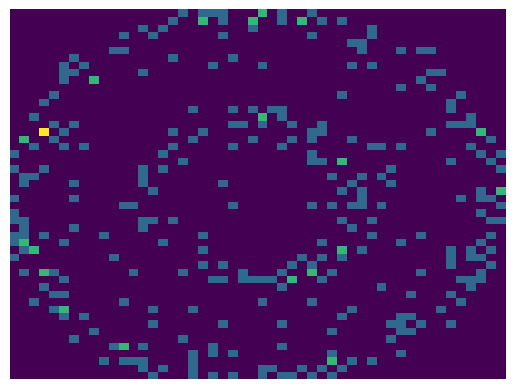}    
		\includegraphics[height=.08\columnwidth, width=.08\columnwidth]{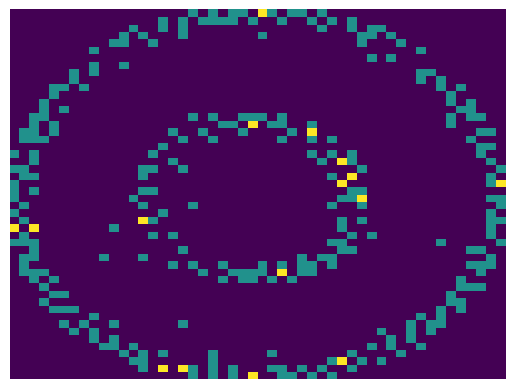}    
		\includegraphics[height=.08\columnwidth, width=.08\columnwidth]{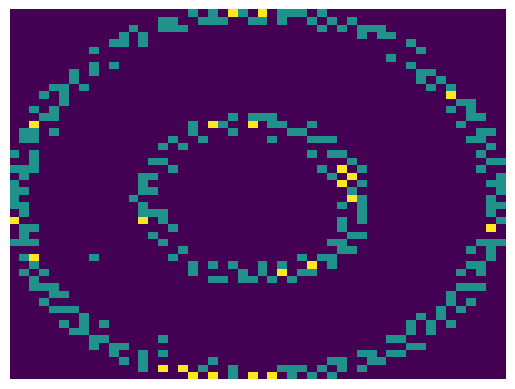}    
		\includegraphics[height=.08\columnwidth, width=.08\columnwidth]{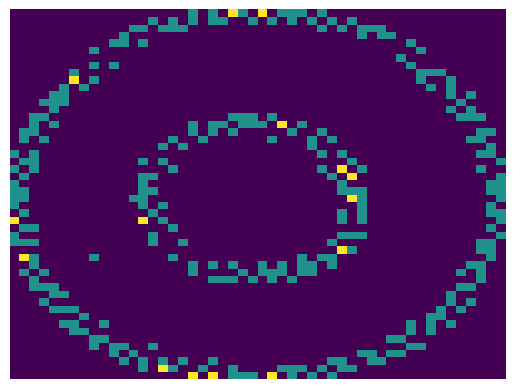}    
		\includegraphics[height=.08\columnwidth, width=.08\columnwidth]{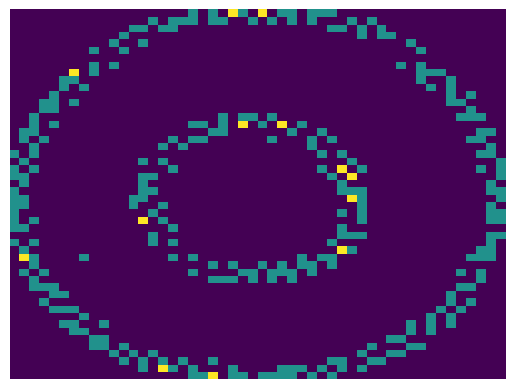} 
		&  
		\includegraphics[height= .08\columnwidth, width=.08\columnwidth]{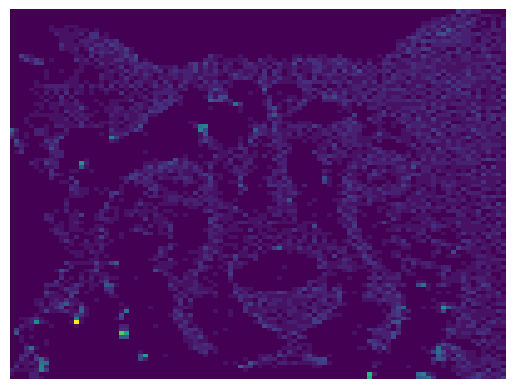}  	
		\includegraphics[height= .08\columnwidth, width=.08\columnwidth]{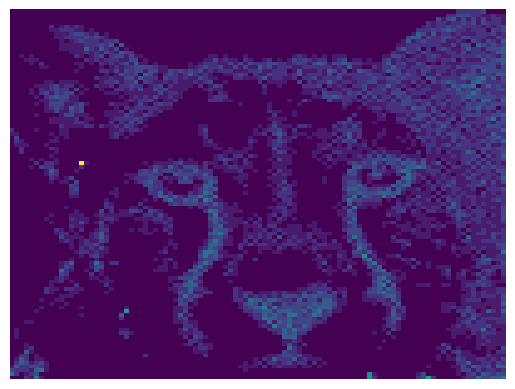} 	
		\includegraphics[height= .08\columnwidth, width=.08\columnwidth]{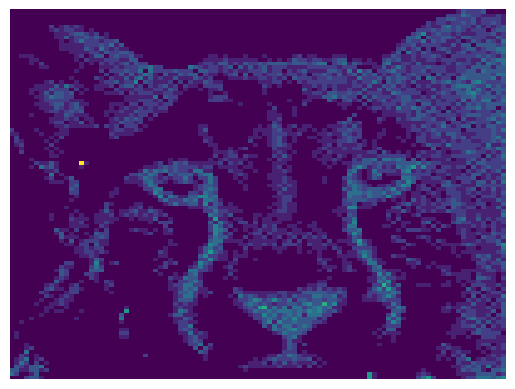} 
		\includegraphics[height= .08\columnwidth, width=.08\columnwidth]{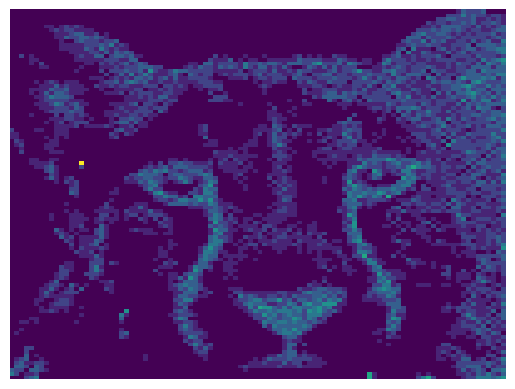} 
		\includegraphics[height= .08\columnwidth, width=.08\columnwidth]{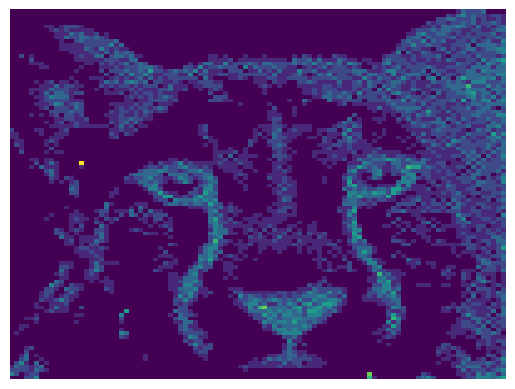} 
		\includegraphics[height= .08\columnwidth, width=.08\columnwidth]{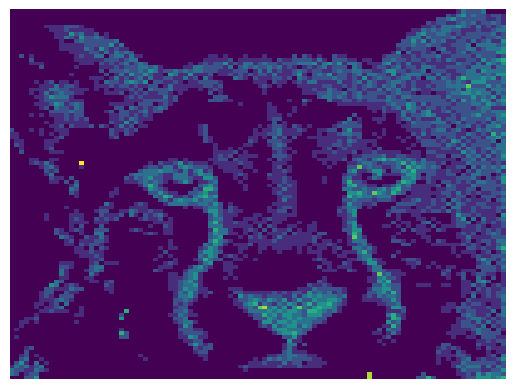}  \\
		($a_1$) \texttt{SD} on ellipses  & 
		($b_1$) \texttt{SD} on sketching\\
		left to right, using 1 to 9 \texttt{SD} steps; &
		left to right, using 1 to 201 \texttt{SD} steps \\
		\includegraphics[height=.08\columnwidth, width=.08\columnwidth]{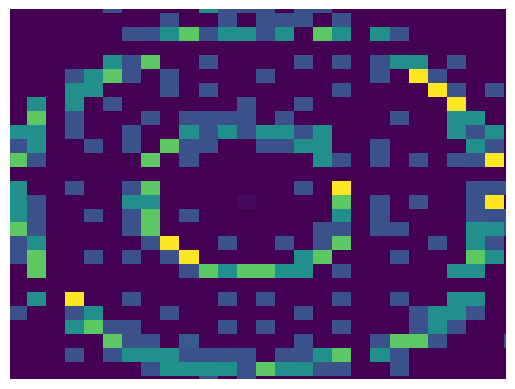}  
		\includegraphics[height=.08\columnwidth, width=.08\columnwidth]{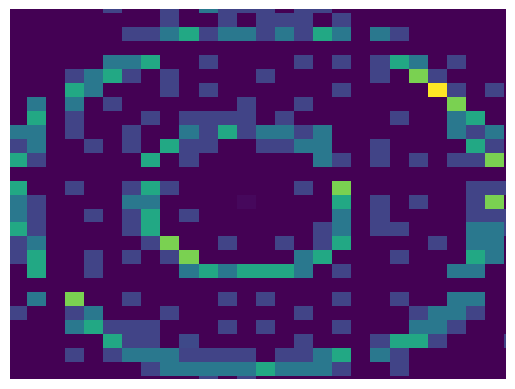}  
		\includegraphics[height=.08\columnwidth, width=.08\columnwidth]{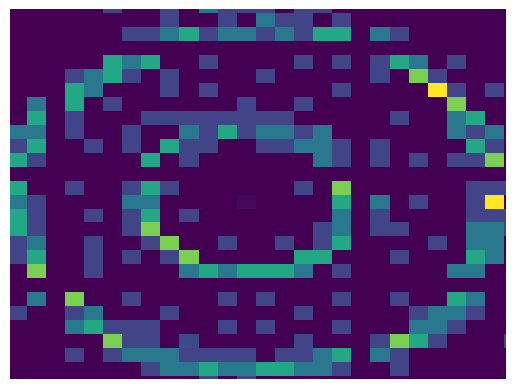}  
		\includegraphics[height=.08\columnwidth, width=.08\columnwidth]{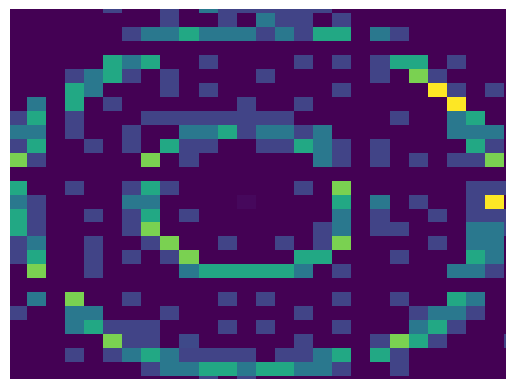} 
		\includegraphics[height=.08\columnwidth, width=.08\columnwidth]{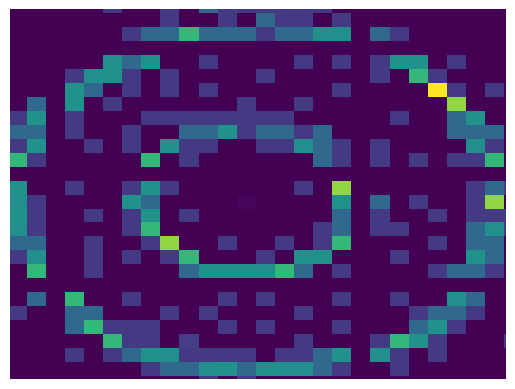} 
		& 
		\includegraphics[height= .08\columnwidth, width=.08\columnwidth]{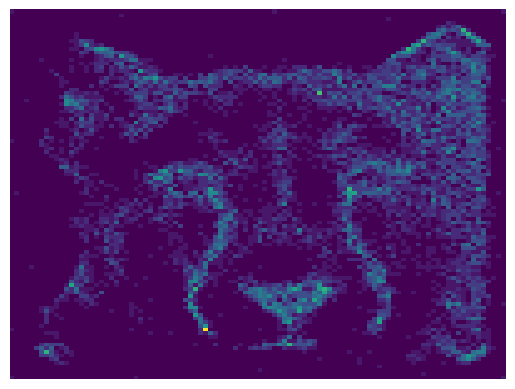} 
		\includegraphics[height= .08\columnwidth, width=.08\columnwidth]{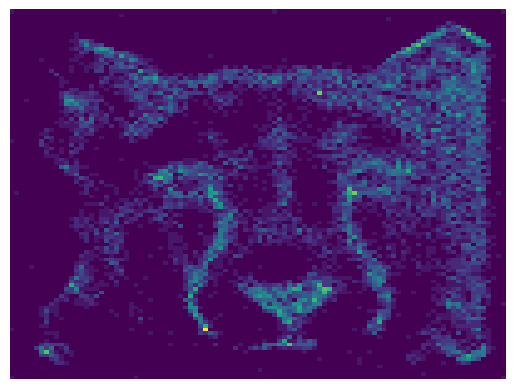} 
		\includegraphics[height= .08\columnwidth, width=.08\columnwidth]{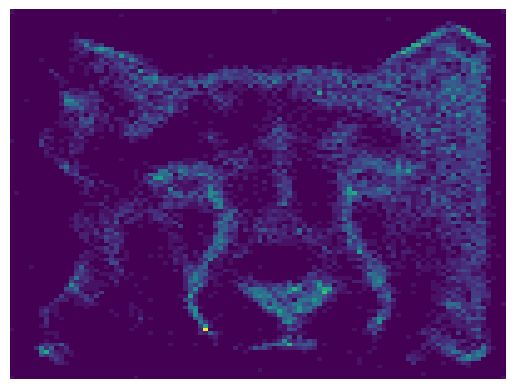} 
		\includegraphics[height= .08\columnwidth, width=.08\columnwidth]{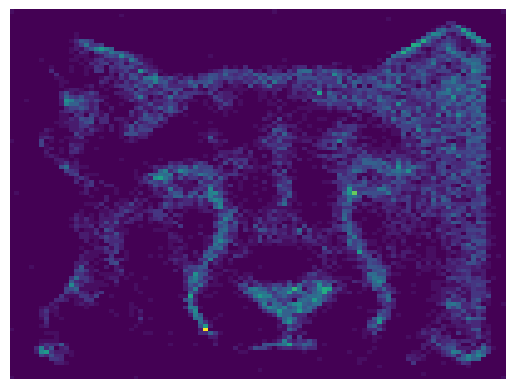} 
		\includegraphics[height= .08\columnwidth, width=.08\columnwidth]{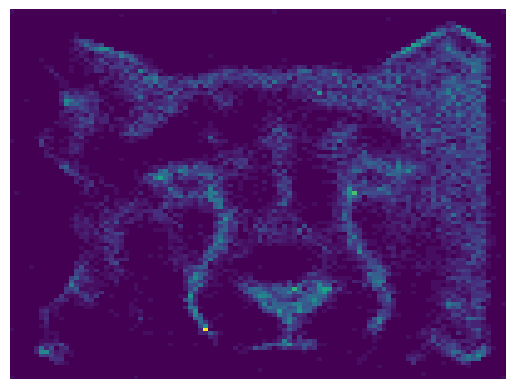} 
		\includegraphics[height= .08\columnwidth, width=.08\columnwidth]{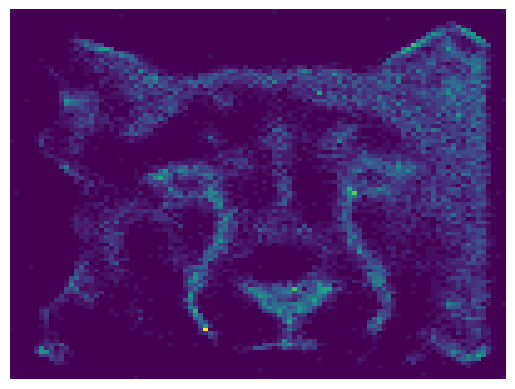} \\
		($a_2$) \texttt{FW} on ellipses  & 
		($b_2$) \texttt{FW} on sketching\\
		left to right, using 411 to 491 \texttt{FW} steps; &
		left to right, using 9901 to 19901 \texttt{FW} steps
	\end{tabular}
	\caption{Visual results of the ellipses and sketching problem.}
	\label{fig_visual}
\end{figure*}

\paragraph{Barycenter of Concentric Ellipses}
We compute the barycenter of 30 randomly generated concentric ellipses similarly as done in \citep{cuturi2014fast,NIPS2019_9130}. 
We run \fw for $500$ iterations and hence the output measure of \fw has support size $N=600$ (\fw increases its support size by $1$ in each iteration). \texttt{SD} is initialized with a discrete uniform distribution with support size varying from $N \in \{20, 40, 80\}$. Note that in these experiments  the chosen support size for \texttt{SD} is even smaller than the initial support size of \fw. 
The result is reported in Figure \ref{fig}(a). 
In terms of convergence rate, we observe that \texttt{SD} is much faster than \fw. Even $20$ iterations are sufficient for \texttt{SD} to find a good solution.
More importantly, in terms of the quality of the solution, \texttt{SD} with support size $N = 20$ outperforms \fw with final support size $N=600$.
In fact, \fw cannot find a solution with better quality even with a larger support size.
This phenomenon is due to an inevitable limitation of the \fw optimization procedure: Each \fw step requires to globally minimize the non-convex function (32) via an exhaustive grid search. This introduces an inherent error to the procedure as the actual solution to (32) potentially resides outside the grid points. Such error limits the accuracy of FW even when the number of particles grows. In contrast, \texttt{SD} adjusts the particles to minimize the objective without any inherent error. As a result, we observe \texttt{SD} outperforms FW on both efficiency and accuracy.
\paragraph{Distribution Sketching}
We consider a special case of the barycenter problem where we only have one source distribution, similarly as done in \citep{NIPS2019_9130}.
This problem can be viewed as approximating a given distribution with a fixed support size budget and is hence called distribution sketching.
Specifically, a natural image of a cheetah is used as the source measure in $\RBB^2$.
We run \fw for $20000$ iterations and  the support size of \texttt{SD} is $N \in \{2000, 4000\}$.
The result is reported in Figure \ref{fig}(b). 
Since we only have one source measure, the Sinkhorn barycenter loss is very small and hence we use log-scale in the y-axis.
We can  observe that \texttt{SD} outperforms \fw in terms of the quality of the solution as well as the convergence rate.
\paragraph{Barycenter of Gaussians}
To demonstrate the efficiency of \texttt{SD} on high dimensional problems, we consider the problem of finding the barycenter of multivariate Gaussian distributions.
Concretely, we pick $5$ isotropic Gaussians in $\RBB^{100}$ with different means.
For each of them, we sample an empirical measure with $50000$ points and used the obtained empirical measures as source measures.
We initialize \texttt{SD} with an empirical measure sampled from the uniform distribution with support size $N = 5000$.
We did not compare with \fw as the global minimizer of $Q(x)$ can not be computed in $\RBB^{100}$.
The result is reported in Figure \ref{fig}(c). 
We can see that just like the previous two experiments, \texttt{SD} converges in less than $20$ iterations.
\paragraph{Visual Results on Ellipses and Sketching.}
To compare \texttt{SD} with \texttt{FW} visually, we allow \texttt{SD} with \texttt{FW} to have a similar amount of particles in the ellipses and sketching tasks, and report the results in Figure \ref{fig_visual}. Specifically, in $(a_1)$ \texttt{SD} has 500 particles while in $(a_2)$ \texttt{FW} has 511 to 591 particles (recall that the support size of \texttt{FW} grows over iterations); in $(b_1)$ \texttt{SD} has 8000 particles while in $(a_2)$ \texttt{FW} has 10001 to 20001 particles.
In all cases \texttt{FW} has at least as much particles as \texttt{SD} does while having significantly more steps.
However, the visual result produced by \texttt{SD} is clearly better than \texttt{FW}: in $(a_1)$, the circle is very clear in the last picture while in $(a_2)$ all pictures remain vague; in $(b_1)$, the eyes of cheetah are clear, but in $(b_2)$ the eyes remain gloomy.
\clearpage
\section{Broader Impact}
This work has the following potential positive impact in the society:
We propose the first algorithm for the Sinkhorn barycenter problem that is scalable with respect to the problem dimension $d$ (linear dependence), while existing works all have an exponential dependence on $d$. 
Further, we expect that this functional gradient descent method can be applied to more general optimization problems involving distribution sampling: In principle, the negative gradient of the dual variables instructs the particles in the measure to search the landscape of the minimizer.

\bibliographystyle{abbrvnat}
\bibliography{SD}

\begin{thebibliography}{25}
\providecommand{\natexlab}[1]{#1}
\providecommand{\url}[1]{\texttt{#1}}
\expandafter\ifx\csname urlstyle\endcsname\relax
  \providecommand{\doi}[1]{doi: #1}\else
  \providecommand{\doi}{doi: \begingroup \urlstyle{rm}\Url}\fi

\bibitem[Ambrosio and Gigli(2013)]{ambrosio2013user}
L.~Ambrosio and N.~Gigli.
\newblock A user’s guide to optimal transport.
\newblock pages 1--155, 2013.

\bibitem[Arbel et~al.(2019)Arbel, Korba, Salim, and Gretton]{arbel2019maximum}
M.~Arbel, A.~Korba, A.~Salim, and A.~Gretton.
\newblock Maximum mean discrepancy gradient flow.
\newblock In \emph{Advances in Neural Information Processing Systems}, pages
  6481--6491, 2019.

\bibitem[Claici et~al.(2018)Claici, Chien, and Solomon]{claici2018stochastic}
S.~Claici, E.~Chien, and J.~Solomon.
\newblock Stochastic wasserstein barycenters.
\newblock In \emph{International Conference on Machine Learning}, pages
  999--1008, 2018.

\bibitem[Cuturi(2013)]{cuturi2013sinkhorn}
M.~Cuturi.
\newblock Sinkhorn distances: Lightspeed computation of optimal transport.
\newblock In \emph{Advances in neural information processing systems}, pages
  2292--2300, 2013.

\bibitem[Cuturi and Doucet(2014)]{cuturi2014fast}
M.~Cuturi and A.~Doucet.
\newblock Fast computation of wasserstein barycenters.
\newblock In \emph{International Conference on Machine Learning}, pages
  685--693, 2014.

\bibitem[Dvurechenskii et~al.(2018)Dvurechenskii, Dvinskikh, Gasnikov, Uribe,
  and Nedich]{dvurechenskii2018decentralize}
P.~Dvurechenskii, D.~Dvinskikh, A.~Gasnikov, C.~Uribe, and A.~Nedich.
\newblock Decentralize and randomize: Faster algorithm for wasserstein
  barycenters.
\newblock In \emph{Advances in Neural Information Processing Systems}, pages
  10760--10770, 2018.

\bibitem[Feydy et~al.(2019)Feydy, S{\'e}journ{\'e}, Vialard, Amari, Trouve, and
  Peyr{\'e}]{feydy2019interpolating}
J.~Feydy, T.~S{\'e}journ{\'e}, F.-X. Vialard, S.-i. Amari, A.~Trouve, and
  G.~Peyr{\'e}.
\newblock Interpolating between optimal transport and mmd using sinkhorn
  divergences.
\newblock In \emph{The 22nd International Conference on Artificial Intelligence
  and Statistics}, pages 2681--2690, 2019.

\bibitem[Genevay et~al.(2016)Genevay, Cuturi, Peyr{\'e}, and
  Bach]{genevay2016stochastic}
A.~Genevay, M.~Cuturi, G.~Peyr{\'e}, and F.~Bach.
\newblock Stochastic optimization for large-scale optimal transport.
\newblock In \emph{Advances in neural information processing systems}, pages
  3440--3448, 2016.

\bibitem[Genevay et~al.(2019{\natexlab{a}})Genevay, Chizat, Bach, Cuturi, and
  Peyr{\'e}]{genevay2019sample}
A.~Genevay, L.~Chizat, F.~Bach, M.~Cuturi, and G.~Peyr{\'e}.
\newblock Sample complexity of sinkhorn divergences.
\newblock In \emph{Proc. AISTATS'19}, 2019{\natexlab{a}}.

\bibitem[Genevay et~al.(2019{\natexlab{b}})Genevay, Chizat, Bach, Cuturi, and
  Peyr\'{e}]{pmlr-v89-genevay19a}
A.~Genevay, L.~Chizat, F.~Bach, M.~Cuturi, and G.~Peyr\'{e}.
\newblock Sample complexity of sinkhorn divergences.
\newblock In K.~Chaudhuri and M.~Sugiyama, editors, \emph{Proceedings of
  Machine Learning Research}, volume~89 of \emph{Proceedings of Machine
  Learning Research}, pages 1574--1583. PMLR, 16--18 Apr 2019{\natexlab{b}}.

\bibitem[Golse(2016)]{golse2016dynamics}
F.~Golse.
\newblock On the dynamics of large particle systems in the mean field limit.
\newblock In \emph{Macroscopic and large scale phenomena: coarse graining, mean
  field limits and ergodicity}, pages 1--144. Springer, 2016.

\bibitem[Kroshnin et~al.(2019)Kroshnin, Tupitsa, Dvinskikh, Dvurechensky,
  Gasnikov, and Uribe]{kroshnin2019complexity}
A.~Kroshnin, N.~Tupitsa, D.~Dvinskikh, P.~Dvurechensky, A.~Gasnikov, and
  C.~Uribe.
\newblock On the complexity of approximating wasserstein barycenters.
\newblock In \emph{International Conference on Machine Learning}, pages
  3530--3540, 2019.

\bibitem[Lemmens and Nussbaum(2012)]{lemmens2012nonlinear}
B.~Lemmens and R.~Nussbaum.
\newblock \emph{Nonlinear Perron-Frobenius Theory}, volume 189.
\newblock Cambridge University Press, 2012.

\bibitem[Liu(2017)]{liu2017stein}
Q.~Liu.
\newblock Stein variational gradient descent as gradient flow.
\newblock In \emph{Advances in neural information processing systems}, pages
  3115--3123, 2017.

\bibitem[Liu and Wang(2016)]{liu2016stein}
Q.~Liu and D.~Wang.
\newblock Stein variational gradient descent: A general purpose bayesian
  inference algorithm.
\newblock In \emph{Advances in neural information processing systems}, pages
  2378--2386, 2016.

\bibitem[Lu et~al.(2019)Lu, Lu, and Nolen]{lu2019scaling}
J.~Lu, Y.~Lu, and J.~Nolen.
\newblock Scaling limit of the stein variational gradient descent: The mean
  field regime.
\newblock \emph{SIAM Journal on Mathematical Analysis}, 51\penalty0
  (2):\penalty0 648--671, 2019.

\bibitem[Luise et~al.(2019)Luise, Salzo, Pontil, and Ciliberto]{NIPS2019_9130}
G.~Luise, S.~Salzo, M.~Pontil, and C.~Ciliberto.
\newblock Sinkhorn barycenters with free support via frank-wolfe algorithm.
\newblock In \emph{Advances in Neural Information Processing Systems 32}. 2019.

\bibitem[Mroueh et~al.(2019)Mroueh, Sercu, and Raj]{mroueh2019sobolev}
Y.~Mroueh, T.~Sercu, and A.~Raj.
\newblock Sobolev descent.
\newblock In \emph{The 22nd International Conference on Artificial Intelligence
  and Statistics}, pages 2976--2985, 2019.

\bibitem[Peyr{\'e} et~al.(2019)Peyr{\'e}, Cuturi,
  et~al.]{peyre2019computational}
G.~Peyr{\'e}, M.~Cuturi, et~al.
\newblock Computational optimal transport.
\newblock \emph{Foundations and Trends{\textregistered} in Machine Learning},
  11\penalty0 (5-6):\penalty0 355--607, 2019.

\bibitem[Rabin et~al.(2011)Rabin, Peyr{\'e}, Delon, and
  Bernot]{rabin2011wasserstein}
J.~Rabin, G.~Peyr{\'e}, J.~Delon, and M.~Bernot.
\newblock Wasserstein barycenter and its application to texture mixing.
\newblock In \emph{International Conference on Scale Space and Variational
  Methods in Computer Vision}, pages 435--446. Springer, 2011.

\bibitem[Solomon et~al.(2015)Solomon, De~Goes, Peyr{\'e}, Cuturi, Butscher,
  Nguyen, Du, and Guibas]{solomon2015convolutional}
J.~Solomon, F.~De~Goes, G.~Peyr{\'e}, M.~Cuturi, A.~Butscher, A.~Nguyen, T.~Du,
  and L.~Guibas.
\newblock Convolutional wasserstein distances: Efficient optimal transportation
  on geometric domains.
\newblock \emph{ACM Transactions on Graphics (TOG)}, 34\penalty0 (4):\penalty0
  1--11, 2015.

\bibitem[Srivastava et~al.(2015)Srivastava, Cevher, Dinh, and
  Dunson]{srivastava2015wasp}
S.~Srivastava, V.~Cevher, Q.~Dinh, and D.~Dunson.
\newblock Wasp: Scalable bayes via barycenters of subset posteriors.
\newblock In \emph{Artificial Intelligence and Statistics}, pages 912--920,
  2015.

\bibitem[Staib et~al.(2017)Staib, Claici, Solomon, and
  Jegelka]{staib2017parallel}
M.~Staib, S.~Claici, J.~M. Solomon, and S.~Jegelka.
\newblock Parallel streaming wasserstein barycenters.
\newblock In \emph{Advances in Neural Information Processing Systems}, pages
  2647--2658, 2017.

\bibitem[Van Der~Vaart and Wellner(1996)]{van1996weak}
A.~W. Van Der~Vaart and J.~A. Wellner.
\newblock Weak convergence.
\newblock In \emph{Weak convergence and empirical processes}, pages 16--28.
  Springer, 1996.

\bibitem[Ye et~al.(2017)Ye, Wu, Wang, and Li]{ye2017fast}
J.~Ye, P.~Wu, J.~Z. Wang, and J.~Li.
\newblock Fast discrete distribution clustering using wasserstein barycenter
  with sparse support.
\newblock \emph{IEEE Transactions on Signal Processing}, 65\penalty0
  (9):\penalty0 2317--2332, 2017.

\end{thebibliography}
\clearpage
\appendix

\section{Preliminaries on the Sinkhorn Potentials}
\begin{lemma}[Lemma \ref{lemma_sinkhorn_potential_bound} elaborated] \label{lemma_optimality_sinkhorn_potential_appendix}
	For a probability measure $\alpha\in\MM_1^+(\XM)$, use $\alpha -a.e.$ to denote ``almost everywhere w.r.t. $\alpha$".
	The pair $(f, g)$ are the Sinkhorn potentials of the entropy-regularized optimal transport problem \eqref{eqn_OTepsilon_dual} if they satisfy 
	\vspace{-.15cm}
	\begin{equation}
		f = \AM(g, \beta), \alpha-a.e. \quad \textrm{and}\quad g = \AM(f, \alpha), \beta-a.e., 
	\end{equation}
	or equivalently 
		\vspace{-.2cm}
	\begin{align}
			\vspace{-.2cm}
		\int_\XM h(x, y)\dB \beta(y) = 1,\ \alpha-a.e., \label{eqn_optimality_sinkhorn_potential_x}\\
				\vspace{-.2cm}
		\int_\XM h(x, y)\dB \alpha(x) = 1,\ \beta-a.e., \label{eqn_optimality_sinkhorn_potential_y}
				\vspace{-.2cm}
	\end{align}
	where $h(x, y) \defi \exp\left(\frac{1}{\gamma}(f(x) + g(y) - c(x, y))\right)$.
\end{lemma}
One can observe that the Sinkhorn potentials are not unique.
In fact, for $\alpha \neq \beta$, the pair $(f_{\alpha,\beta}, g_{\alpha,\beta})$ remains optimal under a constant shift, i.e. $(f_{\alpha,\beta}+C, g_{\alpha,\beta}-C)$ are still the Sinkhorn potentials of $\OTgamma(\alpha, \beta)$ for an arbitrary finite $C\in\RBB$.
Fortunately, it is proved in \cite{cuturi2013sinkhorn} that the Sinkhorn potentials are unique up to such scalar translation.
\\
To reduce the ambiguity, we fix an $x_o \in \XM$ and choose $f_{\alpha,\beta}(x_o) = 0$, since otherwise we can always shift {$f_{\alpha,\beta}$ and $g_{\alpha,\beta}$} by the amount of $f_{\alpha,\beta}(x_o)$.
While it is possible that $x_o\notin\supp(\alpha)$, such choice of $f_{\alpha,\beta}$ is still feasible.
This is because the Sinkhorn potentials can be naturally extended to the entire $\XM$ from Lemma \ref{lemma_optimality_sinkhorn_potential}, even though the above optimality condition characterizes the Sinkhorn potentials on $\supp(\alpha), \supp(\beta)$ only.\\
Further, this choice of $f_{\alpha,\beta}$ allows us to bound $\|f_{\alpha,\beta}\|_\infty$ given that the ground cost function $c$ is bounded on $\XM$.
\begin{assumption}\label{ass_bounded_c}
	The cost function $c(x, y)$ is bounded: $\forall x,y\in\XM, c(x, y)\leq M_c$.
\end{assumption}
\begin{lemma}[Boundedness of the Sinkhorn Potentials] \label{lemma_sinkhorn_potential_bound}
	Let $(f, g)$ be the Sinkhorn potentials of problem \eqref{eqn_OTepsilon_dual} and assume that there exists $x_o\in\XM$ such that $f(x_o) = 0$ (otherwise shift the pair by $f(x_o)$). Then, under Assumption \ref{ass_bounded_c}, $\|f\|_\infty \leq 2M_c$ and $\|g\|_\infty \leq 2M_c$.
\end{lemma}
Next, we analyze the Lipschitz continuity of the Sinkhorn potential $f_{\alpha,\beta}(x)$ with respect to $x$.

\begin{assumption}\label{ass_bounded_infty_c_gradient}
	The cost function $c$ is $G_c$-Lipschitz continuous with respect to one of its inputs: 
	$$\forall x, x' \in \XM, |c(x, y) - c(x', y)|\leq G_c\|x - x'\|.$$
\end{assumption}
Assumption \ref{ass_bounded_infty_c_gradient} implies that $\nabla_x c(x,y)$ exists and for all $x, y\in\XM, \|\nabla_x c(x,y)\|\leq G_c$.
It further ensures the Lipschitz-continuity of the Sinkhorn potential.
\begin{lemma}[Proposition 12 of \cite{feydy2019interpolating}]
	\label{lemma_lipschitz_sinkhorn_potential}
	Under Assumption \ref{ass_bounded_infty_c_gradient}, for a fixed pair of measures $(\alpha, \beta)$, the Sinkhorn potential $f_{\alpha, \beta}:\XM\rightarrow\RBB$ is $G_c$-Lipschitz continuous,
	\begin{equation}
	\forall x, x' \in \XM, |f_{\alpha, \beta}(x) - f_{\alpha, \beta}(x')|\leq G_c\|x - x'\|.
	\end{equation}
	Further, the gradient $\nabla f_{\alpha, \beta}$ exists {at every point $x \in \XM$}, and $\|\nabla f_{\alpha, \beta}(x)\|\leq G_c, \forall x\in\XM$.
\end{lemma}
\begin{assumption}\label{ass_bounded_infty_c_hessian}
	The gradient of the cost function $c$ is $L_c$-Lipschitz continuous: for all $x, x' \in \XM$, $$\|\nabla_1 c(x, y) - \nabla_1 c(x', y)\|\leq L_c\|x - x'\|.$$
\end{assumption}
\begin{lemma}\label{lemma_lipschitz_sinkhorn_potential_gradient}
	Assume Assumptions \ref{ass_bounded_infty_c_gradient} and \ref{ass_bounded_infty_c_hessian}, and denote $L_f \defi {4G_c^2}/{\gamma}+L_c$.
	For a pair of measures $(\alpha, \beta)$, the gradient of the corresponding Sinkhorn potential $f_{\alpha, \beta}:\XM\rightarrow\RBB$ is Lipschitz continuous,
	\begin{equation}
		\forall x, x' \in \XM, \|\nabla f_{\alpha, \beta}(x) - \nabla f_{\alpha, \beta}(x')\|\leq L_f\|x - x'\|.
	\end{equation}
\end{lemma}

\subsection{Computation of Sinkhorn Potentials}\label{section_computation_of_sinkhorn_potential}
The Sinkhorn potential is the cornerstone of the entropy regularized OT problem $\OTgamma(\alpha,\beta)$. Hence, a key component of our method is to efficiently compute this quantity.  
An efficient method is given in \cite{genevay2016stochastic}   when both $\alpha$ and $\beta$ are discrete measures (discrete case), as well as when $\alpha$ is discrete but $\beta$ is continuous (semi-discrete case). More precisely, 
by plugging in the optimality condition on $g$ in \eqref{eqn_optimality_sinkhorn_potential_xy}, the dual problem \eqref{eqn_OTepsilon_dual} becomes
\begin{equation} \label{eqn_regularized_OT_semi_dual}
	\OTgamma(\alpha, \beta) = \max_{f\in\CM} \langle f, \alpha\rangle + \langle\AM(f, \alpha), \beta\rangle.
\end{equation}
Note that \eqref{eqn_regularized_OT_semi_dual} only depends on the values of $f$ on the support of $\alpha$, $\rm{supp}(\alpha)$, which can be represented by a finite dimensional vector $\fB\in\RBB^{|\supp(\alpha)|}$.
Viewing the discrete measure $\alpha$ as a weight vector $\omega_{\alpha}$  on $\supp(\alpha)$,
we have 
\begin{equation*}
	\OTgamma(\alpha, \beta) = \max_{\fB\in\RBB^d} \left\{F(\fB) := \fB^\top\omega_{\alpha} + \EBB_{y\sim\beta}\left[\AM(\fB, \alpha)(y) \right]\right\},
\end{equation*}
that is, $\OTgamma(\alpha, \beta)$ is equivalent to a standard concave stochastic optimization problem, where randomness of the problem comes from $\beta$ (see Proposition 2.1 in \cite{genevay2016stochastic}). 
Hence, the problem can be solved using off-the-shelf stochastic optimization methods. 
In the main body, this method is referred as $\mathcal{SP}_{\gamma}(\alpha,\beta)$.

\section{Lipschitz Continuity of the Sinkhorn Potential}
In this section, we provide several lemmas to show  the Lipschitz continuity (w.r.t. the underlying probability measures) of the Sinkhorn potentials and the functional gradients we derived in Proposition \ref{proposition_variation_I}. 
These lemmas will be used in the convergence analysis and the mean field analysis for \sd.
\subsection{Lipschitz Continuity Study: Sinkhorn Potentials}
We first show the Lipschitz continuity of the Sinkhorn potential w.r.t. the bounded Lipschitz norm of the input measures.
The bounded Lipschitz metric of measures $d_{bl}:\MM_1^+(\XM)\times \MM_1^+(\XM)\rightarrow\RBB_+$ with respect to the bounded continuous test functions is defined as
\begin{equation*}
d_{bl}(\alpha, \beta) \defi \sup_{\|\xi\|_{bl}\leq 1} |\langle \xi, \alpha\rangle - \langle \xi, \beta \rangle|,
\end{equation*}
where, given a function $\xi\in\CM(\XM)$, we denote
\begin{align*}
\|\xi\|_{bl} \defi \max\{\|\xi\|_\infty, \|\xi\|_{lip} \}, \quad \text{with} \|\xi\|_{lip}\defi \max_{x, y\in\XM}\frac{|\xi(x)-\xi(y)|}{\|x-y\|}.
\end{align*}
We note that $d_{bl}$ metrizes the weak convergence of probability measures (see Theorem 1.12.4 in \cite{van1996weak}), i.e. for a sequence of probability measures $\{\alpha_n\}$, $$\lim_{n\rightarrow\infty}d_{bl}(\alpha_n, \alpha) = 0 \Leftrightarrow \alpha_n \rightharpoonup \alpha.$$
\begin{lemma} \label{lemma_Lipschitz_continuity_of_variation}
	(i) Under Assumptions \ref{ass_bounded_c} and \ref{ass_bounded_infty_c_gradient}, for two given pairs of measures $(\alpha, \beta)$ and $(\alpha',\beta')$, the Sinkhorn potentials are Lipschitz continuous with respect to the bounded Lipschitz metric:
	\begin{align*}
	\|f_{\alpha, \beta} - f_{\alpha', \beta'}\|_\infty \leq G_{bl}  [d_{bl}(\alpha', \alpha)+d_{bl}(\beta', \beta)], \\
	\|g_{\alpha, \beta} - g_{\alpha', \beta'}\|_\infty \leq G_{bl}  [d_{bl}(\alpha', \alpha)+d_{bl}(\beta', \beta)].
	\end{align*}
	where $G_{bl} = {2\gamma\exp(2M_c/\gamma)G'_{bl}}/{(1-\lambda^2)}$ with $G'_{bl} = \max\{\exp(3M_c/\gamma), {2G_c\exp(3M_c/\gamma)}/{\gamma}\}$ and $\lambda = \frac{\exp(M_c/\gamma) - 1}{\exp(M_c/\gamma) + 1}$.\\
	(ii) If $(\alpha',\beta')$ are of the particular form $\alpha'={T_\phi}_\sharp\alpha$ and $\beta' = \beta$ where $T_\phi(x) = x + \phi(x), \phi\in\HM^d$, we further have that
	the Sinkhorn potentials are Lipschitz continuous with respect to the mapping $\phi$. That is, letting $G_T \defi {2 G_c\exp(3M_c/\gamma)}/{\gamma}$ and $\epsilon>0$, we have
	\begin{align*}
	\|f_{T_\sharp\alpha, \beta} - f_{\alpha, \beta}\|_\infty \leq  G_{T}\|\phi\|_{2, \infty}, \\
	\|g_{T_\sharp\alpha, \beta} - g_{\alpha, \beta}\|_\infty \leq  G_{T}\|\phi\|_{2, \infty}.
	\end{align*}
\end{lemma}
Please see the proof in Appendix \ref{proof_lemma_Lipschitz_continuity_of_variation}.
Importantly, this lemma implies that the weak convergence of $(\alpha,\beta)$ ensures the convergence of the Sinkhorn potential: $(\alpha', \beta')\rightharpoonup(\alpha, \beta)\Rightarrow (f_{\alpha',\beta'}\rightarrow f_{\alpha,\beta})$ in terms of the $L^\infty$ norm.
\begin{remark}
	While we acknowledge that the factor $\exp{1/\gamma}$ is non-ideal, such quantity constantly appears in the literature related to the Sinkhorn divergence, e.g. Theorem 5 in \cite{NIPS2019_9130} and Theorem 3 in \cite{pmlr-v89-genevay19a}. It would be an interesting future work to remove this factor.
\end{remark}
\begin{remark}
	We note that the Lemma \ref{lemma_Lipschitz_continuity_of_variation} is strictly stronger than preexisting results:
	(1) Proposition 13 of \cite{feydy2019interpolating} only shows that the dual potentials are continuous (not Lipschitz continuous) with the input measures, which is insufficient for the mean field limit analysis conducted in Section \ref{section_mean_field_limit}.
	(2) Under the infinity norm $\|\cdot\|_\infty$, \citet{NIPS2019_9130} bound the variation of the Sinkhorn potential by the total variation distance of probability measures $(\alpha, \beta)$ and $(\alpha',\beta')$.
	Such result means that strong convergence of $(\alpha,\beta)$ implies the convergence of the corresponding Sinkhorn potential.
	This is strictly weaker than (i) of Lemma \ref{lemma_Lipschitz_continuity_of_variation}.
	(3) Further, to prove the weak convergence of the corresponding Sinkhorn potential, Proposition E.5 of the above work \cite{NIPS2019_9130} requires the cost function $c \in \CM^{s+1}$ with $s>d/2$, where $d$ is the problem dimension.
	However, Lemma \ref{lemma_Lipschitz_continuity_of_variation} only assumes $c \in \CM^{1}$, independent of $d$.
	Hence, Lemma \ref{lemma_Lipschitz_continuity_of_variation} makes a good contribution over existing results.
\end{remark}
The continuity results in Lemma \ref{lemma_Lipschitz_continuity_of_variation} can be further extended to the gradient of the Sinkhorn potentials.
\begin{lemma} \label{lemma_Lipschitz_continuity_of_variation_gradient}
	(i) Under Assumptions \ref{ass_bounded_c} and \ref{ass_bounded_infty_c_gradient}, for two given pairs of measures $(\alpha, \beta)$ and $(\alpha',\beta')$, with $G_{bl}  [d_{bl}(\alpha', \alpha)+d_{bl}(\beta', \beta)] \leq 1$, the gradient of the Sinkhorn potentials are locally Lipschitz continuous with respect to the bounded Lipschitz metric: With $L_{bl} = 2G_cG_{bl}$,
	\begin{align*}
	\|\nabla f_{\alpha, \beta} - \nabla f_{\alpha', \beta'}\|_\infty \leq L_{bl}  [d_{bl}(\alpha', \alpha)+d_{bl}(\beta', \beta)], \\
	\|\nabla g_{\alpha, \beta} - \nabla g_{\alpha', \beta'}\|_\infty \leq L_{bl}  [d_{bl}(\alpha', \alpha)+d_{bl}(\beta', \beta)].
	\end{align*}
	(ii) If $(\alpha',\beta')$ are of the particular form $\alpha'={T_\phi}_\sharp\alpha$ and $\beta' = \beta$ where $T_\phi(x) = x + \phi(x)$ for $\phi\in\HM^d$, we further have that
	the Sinkhorn potentials are Lipschitz continuous with respect to the mapping $\phi$: Let $G_T \defi {2 G_c\exp(3M_c/\gamma)}/{\gamma}$ and assume $2 G_T\|\phi\|_{2,\infty}\leq 1$. We have with $L_{T} = 2G_cG_T$
	\begin{align*}
	\|\nabla f_{T_\sharp\alpha, \beta} - \nabla f_{\alpha, \beta}\|_\infty \leq  L_{T}\|\phi\|_{2, \infty}, \\
	\|\nabla g_{T_\sharp\alpha, \beta} - \nabla g_{\alpha, \beta}\|_\infty \leq  L_{T}\|\phi\|_{2, \infty}.
	\end{align*}
\end{lemma}
The proof is given in Appendix \ref{proof_lemma_Lipschitz_continuity_of_variation_gradient}. The two lemmas \ref{lemma_Lipschitz_continuity_of_variation} \ref{lemma_Lipschitz_continuity_of_variation_gradient} are crucial to the analysis of the finite-time convergence and the mean field limit of \sinkhorndescent.

\subsection{Lipschitz Continuity Study: \FDcamel}
\vspace{-.2cm}
From Definition \ref{definition_variation_rkhs}, the \FDs derived in Proposition \ref{proposition_variation_I} are functions in $\HM^d$ mapping from $\XM$ to $\RBB^d$. They are Lipschitz continuous provided that the kernel function $k$ is Lipschitz.
\begin{assumption} \label{ass_lipschitz_kernel}
	The kernel function $k:\XM\times\XM\rightarrow\RBB_+$ is Lipschitz continuous on $\XM$: for any $y$ and $x, x'\in\XM$
	\begin{equation}
	|k(x, y) - k(x', y)|\leq G_k\|x - x'\|.
	\end{equation}
\end{assumption}
\begin{lemma} \label{lemma_lipschitz_variation_mapping}
	Define the functional on RKHS $F[\psi]\defi\OTgamma\big({(\IM+\psi)}_\sharp\alpha, \beta\big)$.
	Assume Assumptions \ref{ass_bounded_c}, \ref{ass_bounded_infty_c_gradient}, \ref{ass_bounded_infty_c_hessian}, and \ref{ass_lipschitz_kernel}.
	The \FD $DF[0]\in\HM^d$ is Lipschitz continuous: Denote $L_\psi = G_cG_k$. For any $x, x'\in\XM$,
	\begin{equation*}
	\|D F[0](x) - D F[0](x')\|\leq L_\psi\|x - x'\|.
	\end{equation*}
\end{lemma}
Using the above result, the functional gradient \eqref{eqn_gradient_sinkhorn_barycenter} can be shown to be Lipschitz continuous.
\begin{corollary} \label{corollary_lipschitz_variation_mapping}
	Assume Assumptions \ref{ass_bounded_c}, \ref{ass_bounded_infty_c_gradient}, \ref{ass_bounded_infty_c_hessian}, and \ref{ass_lipschitz_kernel}.
	Recall $L_\psi = G_cG_k$ from the above lemma. 
	The \FD $DS_{\alpha}[0]\in\HM^d$ is Lipschitz continuous: For any $x, x'\in\XM$,
	\begin{equation*}
	\|DS_{\alpha}[0](x) - DS_{\alpha}[0](x')\|\leq L_\psi\|x - x'\|.
	\end{equation*}
\end{corollary}
\subsection{Last term convergence of \sd} \label{appendix_last_term_convergence}
With a slight change to \sd, we can claim its last term convergence: In each iteration, check if $\SB(\alpha^t, \{\beta_i\}_{i=1}^n)\leq \epsilon$. If it holds, then we have already identified an $\epsilon$ approximate stationary point and we terminate \sd; otherwise we proceed. The termination happens within $\OM(1/\epsilon)$ loops as the nonnegative objective \eqref{eqn_sinkhorn_barycenter} is reduced at least $\OM(\epsilon)$ per-round.
\section{Proof of Lemmas}
\subsection{Proof of Lemma \ref{lemma_lipschitz_sinkhorn_potential}} \label{proof_lemma_lipschitz_sinkhorn_potential}
	For simplicity, we omit the subscript of the Sinkhorn potential $f_{\alpha,\beta}$ and simply use $f$.
	Recall the definition of $h(x, y)$ in Lemma \ref{lemma_optimality_sinkhorn_potential_appendix}:
	\begin{equation*}
		h(x, y) = \exp\left(\frac{1}{\gamma}(f(x) + g(y) - c(x, y))\right).
	\end{equation*}
	Subtract the optimality condition \eqref{eqn_optimality_sinkhorn_potential_x} at different points $x$ and $x'$ to derive
	\begin{align*}
		\int_\XM \big(h(x, y) - h(x',y)\big) \dB\beta(y) = 0 \Rightarrow\\
		\int_\XM h(x', y) &\left(\exp(\frac{f(x) - f(x') - c(x, y) + c(x', y)}{\gamma})-1\right) \dB\beta(y) = 0
	\end{align*}
	Since $\int_\XM h(x', y) \dB\beta(y) = 1$ (Lemma \ref{lemma_optimality_sinkhorn_potential_appendix}), we have
	\begin{align*}
		\int_\XM h(x', y) \exp(\frac{f(x) - f(x') - (c(x, y) - c(x', y))}{\gamma}) \dB\beta(y) = 1 \\
		\Rightarrow \int_\XM h(x', y) \exp(\frac{c(x', y) - c(x, y)}{\gamma}) \dB\beta(y) &= \exp(\frac{f(x') - f(x)}{\gamma}).
	\end{align*}
	Further, since we have $h(x',y)\geq0$ and from Assumption \ref{ass_bounded_c} we have$$\exp(\frac{c(x', y) - c(x, y)}{\gamma}) \leq \exp(\frac{|c(x', y) - c(x, y)|}{\gamma}) \leq \exp(\frac{G_c\|x' - x\|}{\gamma}),$$  we derive
	\begin{equation*}
		|\frac{f(x') - f(x)}{\gamma}| \leq |\log(\int_\XM h(x', y) \exp(\frac{G_c\|x' - x\|}{\gamma}) \dB\beta(y))| \leq \frac{G_c\|x' - x\|}{\gamma},
	\end{equation*}
	by using $\int_\XM h(x', y) \dB\beta(y) = 1$ again, which consequently leads to 
	\begin{equation*}
		|f(x') - f(x)|\leq G_c\|x' - x\|.
	\end{equation*}
\subsection{Proof of Lemma \ref{lemma_lipschitz_sinkhorn_potential_gradient}} \label{proof_lemma_lipschitz_sinkhorn_potential_gradient}
Recall the expression of $\nabla f$ in \eqref{eqn_sinkhorn_potential_gradient_x}:
\begin{align}
\nabla f(x) = \int_\XM h(x, y)\nabla_x c(x, y) \dB\beta(y),
\end{align}
where $h(x, y) \defi \exp\left(\frac{1}{\gamma}(f_{\alpha,\beta}(x) + \AM[f_{\alpha,\beta}, \alpha](y) - c(x, y))\right)$.
For any $x, x'\in\XM$ such that $\|x_1 - x_2\|\leq\frac{\gamma}{2G_c}$, we bound
\begin{align*}
	\|\nabla f(x) - \nabla f(x')\| =&\ \|\int_\XM h(x, y)\nabla_x c(x, y) - h(x', y)\nabla_x c(x', y) \dB\beta(y)\| \\
	\leq&\ \int_\XM \|h(x, y)\nabla_x c(x, y) - h(x', y)\nabla_x c(x', y)\| \dB\beta(y)
\end{align*}
To bound the last integral, observe that
\begin{align*}
	h(x, y)\nabla_x c(x, y) - h(x', y)\nabla_x c(x', y)\\
 = h(x, y)\big(\nabla_x c(x, y) - \nabla_x c(x', y)\big) &+ \big(h(x, y) - h(x', y)\big)\nabla_x c(x', y),
\end{align*}
and therefore
\begin{align*}
	\|h(x, y)\nabla_x c(x, y) - h(x', y)\nabla_x c(x', y)\|\\
	\leq h(x, y)\|\nabla_x c(x, y) -\nabla_x c(x', y)\|& + |h(x, y) - h(x', y)|\|\nabla_x c(x', y)\|.
\end{align*}
For the first term, we use the Lipschitz continuity of $\nabla_x c$ from Assumption \ref{ass_bounded_infty_c_hessian} to bound
\begin{equation*}
	h(x, y)\|\nabla_x c(x, y) -\nabla_x c(x', y)\| \leq L_c h(x, y)\|x-x'\|.
\end{equation*}
For the second term, observe that $\|\nabla_x c(x', y)\|\leq G_c$ from Assumption \ref{ass_bounded_infty_c_gradient} and 
\begin{align*}
	|h(x, y) - h(x', y)| =&\ h(x', y)|\exp(\frac{f(x) - f(x') - c(x, y) + c(x', y)}{\gamma}) - 1|\\
	<&\ 2h(x', y)|\frac{f(x) - f(x') - c(x, y) + c(x', y)}{\gamma}|.
\end{align*}
Since $|\exp(z)-1|< 2|z|$ when $|z|\leq 1$ ($z=|\frac{f(x) - f(x') - c(x, y) + c(x', y)}{\gamma}|\leq 1$ from the restriction on $\|x-x'\|$), we further derive
\begin{equation*}
	|h(x, y) - h(x', y)|\leq \frac{2G_c}{\gamma}h(x', y)[2G_c\|x-x'\|] = \frac{4G_c^2}{\gamma}h(x', y) \|x-x'\|.
\end{equation*}
Using the optimality condition $\int_\XM h(x', y) \dB\beta(y) = 1$ and $\int_\XM h(x, y) \dB\beta(y) = 1$ from Lemma \ref{lemma_optimality_sinkhorn_potential}, we derive
\begin{equation*}
	\|\nabla f(x) - \nabla f(x')\|\leq \int_\XM L_c h(x, y)\|x-x'\| + \frac{4G_c^2}{\gamma}h(x', y) \|x-x'\| \dB\beta(y) = (L_c+\frac{4G_c^2}{\gamma})\|x-x'\|.
\end{equation*}
This implies that $\nabla^2 f(x)$ exists and is bounded from above: $\forall x\in\XM, \|\nabla^2 f(x)\|\leq L_f$, which concludes the proof.
\subsection{Proof of Lemma \ref{lemma_Lipschitz_continuity_of_variation}}
\label{proof_lemma_Lipschitz_continuity_of_variation}
Let $(f, g)$ and $(f', g')$ be the Sinkhorn potentials to $\OTgamma(\alpha, \beta)$ and $\OTgamma(\alpha', \beta')$ respectively.
Denote $u \defi \exp(f/\gamma)$, $v \defi \exp(g/\gamma)$ and $u' \defi \exp(f'/\gamma)$, $v' \defi \exp(g'/\gamma)$.
From Lemma \ref{lemma_sinkhorn_potential_bound}, $u$ is bounded in terms of the $L^\infty$ norm:
\begin{equation*}
	\|u\|_\infty = \max_{x\in\XM} |u(x)| = \max_{x\in\XM} \exp(f/\gamma) \leq \exp(2M_c/\gamma),
\end{equation*} 
which also holds for $v, u', v'$.
Additionally, from Lemma \ref{lemma_lipschitz_sinkhorn_potential}, $\nabla u$ exists and $\|\nabla u\|$ is bounded:
\begin{equation*}
\max_x \|\nabla u(x)\| = 	\max_x \frac{1}{\gamma}|u(x)|\|\nabla f(x)\|\leq \frac{1}{\gamma}\|u(x)\|_\infty\max_x\|\nabla f(x)\|\leq 
{G_c\exp(2M_c/\gamma)}/{\gamma}.
\end{equation*}
Define the mapping $A_{\alpha} \mu \defi 1/(L_\alpha \mu)$ with 
\begin{equation*}
L_\alpha \mu = \int_\XM l(\cdot, y)\mu(y)\dB \alpha(y),
\end{equation*}
where $l(x, y) \defi \exp(-c(x, y)/\gamma)$. 
From Assumption \ref{ass_bounded_c}, we have $\|l\|_\infty\leq\exp(M_c/\gamma)$ and from Assumption \ref{ass_bounded_infty_c_gradient} we have $\|\nabla_x l(x, y)\|\leq \exp(M_c/\gamma)\frac{G_c}{\gamma}$.
From the optimality condition of $f$ and $g$, we have $v = A_{\alpha} u$ and $u = A_{\beta} v$. Similarly, $v' = A_{\alpha'} u'$ and $u' = A_{\beta'} v'$.
Further use $d_H:\CM(\XM)\times\CM(\XM)\rightarrow\RBB$ to denote the Hilbert metric of continuous functions, $$d_H(\mu, \nu) = \log \max_{x,x'\in\XM}\frac{\mu(x)\nu(x')}{\mu(x')\nu(x)}.$$
Note that $d_H(\mu, \nu) = d_H(1/\mu, 1/\nu)$ if $\mu(x)>0$ and $\nu(x)>0$ $\forall x\in\XM$ and hence $d_H(L_\alpha\mu, L_\alpha\nu) = d_H(A_\alpha\mu, A_\alpha\nu)$.
Under the above notations, we introduce the following existing result.
\begin{lemma}[Birkhoff-Hopf Theorem \cite{lemmens2012nonlinear}, see Lemma B.4 in \cite{NIPS2019_9130}] 
	\label{lemma_Birkhoff-Hopf}
	Let $\lambda = \frac{\exp(M_c/\gamma) - 1}{\exp(M_c/\gamma) + 1}$ and $\alpha\in\MM_1^+(\XM)$. Then for every $u, v\in\CM(\XM)$, such that $u(x)>0, v(x)>0$ for all $x\in\XM$, we have
	\begin{equation*}
		d_H(L_\alpha u, L_\alpha v)\leq \lambda d_H(u, v).
	\end{equation*}
\end{lemma}
Note that from the definition of $d_H$, one has
\begin{align*}
\|\log\mu-\log\nu \|_\infty\leq d_H(\mu, \nu) =&\ \max_x[\log\mu(x) - \log \nu(x)]+\max_x[\log\nu(x) - \log \mu(x)]\\
\leq&\ 2\|\log\mu-\log\nu \|_\infty.
\end{align*}
In the following, we derive upper bound for $d_H(\mu, \nu)$ and use such bound to analyze the Lipschitz continuity of the Sinkhorn potentials $f$ and $g$.\\
Construct $\tilde{v} \defi A_{\alpha} u'$.
Using the triangle inequality (which holds since $v(x), v'(x), \tilde{v}(x) >0$ for all $x\in\XM$), we have
\begin{align*}
d_H(v, v')\leq d_H(v, \tilde{v}) + d_H(\tilde{v}, v') \leq
\lambda d_H(u, u') + d_H(\tilde{v}, v'),
\end{align*}
where the second inequality is due to Lemma \ref{lemma_Birkhoff-Hopf}.
Similarly, Construct $\tilde{u} \defi A_{\beta} v'$.
Apply Lemma \ref{lemma_Birkhoff-Hopf} again to obtain
\begin{equation*}
	d_H(u, u') \leq d_H(u, \tilde u) + d_H(\tilde u, u')\leq \lambda d_H(v, v') + d_H(\tilde{u}, u').
\end{equation*}
Together, we obtain 
\begin{equation*}
d_H(v, v') \leq \lambda^2d_H(v, v') + d_H(\tilde{v}, v') + \lambda d_H(\tilde{u}, u') \leq \lambda^2d_H(v, v') + d_H(\tilde{v}, v') + d_H(\tilde{u}, u'),
\end{equation*}
which leads to
\begin{equation*}
d_H(v, v') \leq \frac{1}{1- \lambda^2}[d_H(\tilde{v}, v') + d_H(\tilde{u}, u')].
\end{equation*}

To bound $d_H(\tilde{v}, v')$ and similarly $d_H(\tilde{u}, u')$, observe the following:
\begin{align}
d_H(v', \tilde v) =& d_H(L_{\alpha'} u', L_{\alpha} u') \leq 2\|\log L_{\alpha'} u' - \log L_{\alpha} u'\|_\infty \notag\\
=& 2\max_{x\in\XM}| \nabla \log(a_x) ([L_{\alpha'} u'](x) - [L_{\alpha} u'](x))| = 2\max_{x\in\XM} \frac{1}{a_x} |[L_{\alpha'} u'](x) - [L_{\alpha} u'](x)|\notag\\
\leq& 2\max\{\|1/L_{\alpha'} u'\|_\infty, \|1/L_{\alpha} u'\|_\infty\}\|L_{\alpha'} u' - L_{\alpha} u'\|_\infty \label{appendix_proof_i},
\end{align}
where $a_x\in[[L_{\alpha'} u'](x), [L_{\alpha} u'](x)]]$ in the second line is from the mean value theorem.
Further, in the inequality we use $\max\{\|1/L_{\alpha} u'\|_\infty, \|1/L_{\alpha} u'\|_\infty\} = \max\{\|A_{\alpha'} u'\|_\infty, \|A_{\alpha} u'\|_\infty\} \leq \exp(2M_c/\gamma)$.
Consequently, all we need to bound is the last term $\|L_{\alpha'} u' - L_{\alpha} u'\|_\infty$.

{\bf Result (i)} 
We first note that $\forall x\in\XM$, $\|l(x, \cdot)u'(\cdot)\|_{bl}<\infty$: In terms of $\|\cdot\|_\infty$
\begin{equation*}
	\|l(x, \cdot)u'(\cdot)\|_\infty \leq \|l(x, \cdot)\|_\infty\|u'\|_\infty\leq \exp(3M_c/\gamma) <\infty.
\end{equation*}
In terms of $\|\cdot\|_{lip}$, we bound
\begin{align*}
	\|l(x, \cdot)u'(\cdot)\|_{lip} \leq&\ \|l(x, \cdot)\|_\infty\|u'\|_{lip} + \|l(x, \cdot)\|_{lip}\|u'\|_{\infty}\\
	\leq&\ \exp(M_c/\gamma){G_c\exp(2M_c/\gamma)}/{\gamma} + \exp(M_c/\gamma){G_c}\exp(2M_c/\gamma)/{\gamma} \\
	=&\ {2G_c\exp(3M_c/\gamma)}/{\gamma}< \infty.
\end{align*}	
Together we have $\|l(x, y)u'(y)\|_{bl} \leq \max\{\exp(3M_c/\gamma), {2G_c\exp(3M_c/\gamma)}/{\gamma}\}$.
From the definition of the operator $L_{\alpha}$, we have
\begin{align*}
	\|L_{\alpha'} u' - L_{\alpha} u'\|_\infty =&\ \max_x |\int_\XM l(x, y)u'(y)\dB\alpha'(y) - \int_\XM l(x, y)u'(y)\dB\alpha(y)|\\
	\leq&\ \|l(x, y)u'(y)\|_{bl} d_{bl}(\alpha', \alpha).
\end{align*}
All together we derive
\begin{equation*}
	d_H(v', v) \leq \frac{2\exp(2M_c/\gamma)\|l(x, y)u'(y)\|_{bl}}{1-\lambda^2}  [d_{bl}(\alpha', \alpha)+d_{bl}(\beta', \beta)] \quad(\lambda = \frac{\exp(M_c/\gamma) - 1}{\exp(M_c/\gamma) + 1}).
\end{equation*}
Further, since $d_H(v', v) \geq \|\log v'-\log v \|_\infty =  \frac{1}{\gamma}\|f'- f \|_\infty$, we have the result:
\begin{equation}
	\|f'- f \|_\infty\leq \frac{2\gamma\exp(2M_c/\gamma)\|l(x, y)u'(y)\|_{bl}}{1-\lambda^2}  [d_{bl}(\alpha', \alpha)+d_{bl}(\beta', \beta)].
\end{equation}
Similar argument can be made for $\|g'- g \|_\infty$.

{\bf Result (ii)} Recall that $\alpha' = T_\phi\sharp\alpha$ and $\beta' = \beta$ with $T_\phi(x) = x + \phi(x)$.
For simplicity we denote $f' = f_{T_\phi\sharp\alpha, \beta}$ and $g' = g_{T_\phi\sharp\alpha, \beta}$ and $f = f_{\alpha, \beta}$ and $g = g_{\alpha, \beta}$. We denote similarly $u'$, $v'$, $u$, and $v$.
Use \eqref{appendix_proof_i} and the change-of-variables formula of the push-forward measure to obtain
\begin{align*}
\|L_{T_\phi\sharp\alpha} u' - L_{\alpha} u'\|_\infty = \max_x \int [l(x,T_\phi( y))u'(T_\phi(y)) - l(x, y)u'(y)]\dB\alpha(y).
\end{align*}
We now bound the integrand:
\begin{align*}
&|l(x,T_\phi( y))u'(T_\phi(y)) - l(x, y)u'(y)|\\
= &|l(x,T_\phi(y))u'(T_\phi(y)) - l(x,T_\phi( y))u'(y)| + |l(x, T_\phi( y))u'(y) - l(x, y)u'(y)| \\
\leq& \exp(M_c/\gamma)\cdot \frac{G_c\exp(2M_c/\gamma)}{\gamma}\|\phi(y)\| + \exp(M_c/\gamma)\frac{G_c}{\gamma} \cdot\exp(2M_c/\gamma)\cdot\|\phi(y)\|\\
\leq &  \frac{2 G_c\exp(3M_c/\gamma)}{\gamma}\cdot\|\phi(y)\|,
\end{align*}
where we use the Lipschitz continuity of $u'$ for the first term and the Lipschitz continuity of $l$ for the second term.
\subsection{Proof of Lemma \ref{lemma_Lipschitz_continuity_of_variation_gradient}}
\label{proof_lemma_Lipschitz_continuity_of_variation_gradient}
From the restriction on $d_{bl}(\alpha', \alpha)+d_{bl}(\beta', \beta)$ or the size of the mapping $\|\phi\|_\infty$, we always have $|f(x) + g(y) - f'(x) - g'(y)|<1$ from Lemma \ref{lemma_Lipschitz_continuity_of_variation}.\\
Denote the Sinkhorn potentials to $\OTgamma(\alpha, \beta)$ and $\OTgamma(\alpha', \beta')$ by $(f, g)$ and $(f', g')$ respectively.
From the expression \eqref{eqn_sinkhorn_potential_gradient_x} of $\nabla f$ (and $\nabla f'$), we have
\begin{align*}
	\|\nabla f(x) - \nabla f'(x)\| =& \|\int_\XM (h(x, y) - h'(x, y))\nabla_x c(x, y) \dB\beta(y)\|\\
	=& \|\int_\XM h'(x, y)(\exp(f(x) + g(y) - f'(x) - g'(y))-1)\nabla_x c(x, y) \dB\beta(y)\|\\
	\leq& \int_\XM h'(x, y)|\exp(f(x) + g(y) - f'(x) - g'(y))-1|\|\nabla_x c(x, y)\| \dB\beta(y) \\
	\leq& \int_\XM 2h'(x, y)|f(x) + g(y) - f'(x) - g'(y)| \|\nabla_x c(x, y)\| \dB\beta(y),
\end{align*}
where $h'(x, y) \defi \exp(\frac{1}{\gamma}(f'(x) + g'(y) - c(x, y)))$, the second inequality holds since $|exp(x) - 1| < 2|x|$ when $|x|\leq 1$ and $|f(x) + g(y) - f'(x) - g'(y)|<1$. We can use results from Lemma \ref{lemma_Lipschitz_continuity_of_variation} to bound the term $|f(x) + g(y) - f'(x) - g'(y)|$.

\noindent{\bf Result (i):} Using (i) of Lemma \ref{lemma_Lipschitz_continuity_of_variation}, we bound
\begin{equation*}
	\|\nabla f(x) - \nabla f'(x)\| \leq 2G_cG_{bl}[d_{bl}(\alpha', \alpha)+d_{bl}(\beta', \beta)].
\end{equation*}

\noindent{\bf Result (ii):} Using (ii) of Lemma \ref{lemma_Lipschitz_continuity_of_variation}, we bound
\begin{equation*}
	\|\nabla f(x) - \nabla f'(x)\| \leq 2 G_cG_{T}\|\phi\|_\infty.
\end{equation*}
\subsection{Proof of Proposition \ref{proposition_variation_I}}
\label{proof_proposition_variation_I}
	We will compute $DF_1[0]$ based on the definition of the Fr\'echet derivatives in Definition \ref{definition_variation_rkhs}.
	The computation of $DF_2[0]$ follows similarly.\\
	Denote $T_\psi = \IM + \psi$.
	Note that we are interested in the case when $\psi=0$ and hence $T_{\psi+\epsilon\phi}(x) = T_{\epsilon\phi}(x) = x + \epsilon\phi(x)$.
	Additionally, $T_\psi$ is the identity operator when $\psi = 0$ and hence $F_1[0] = \OTgamma(\alpha, \beta)$.
	For simplicity, we drop the subscript of $T_{\epsilon\phi}$ ($\psi = 0$) and simply denote it by $T$ in the rest of the proof.
	Let $f$ and $g$ be the Sinkhorn potentials to $\OTgamma(\alpha, \beta)$, by \eqref{eqn_OTepsilon_dual} and the optimality of $f$ and $g$, one has 
	\[
	\OTgamma(\alpha, \beta) = \langle f, \alpha \rangle + \langle g, \beta  \rangle. 
	\]	
	However, $f$ and $g$ are not necessarily the optimal dual variables for $\OTgamma(T_\sharp\alpha, \beta)$, so one has 	
	\[
	\OTgamma(T_\sharp\alpha, \beta) \geq \langle f, T_\sharp\alpha \rangle + \langle g, \beta\rangle - \gamma \langle h-1, T_\sharp\alpha\otimes\beta \rangle. 
	\]
	Using the optimality from Lemma \ref{lemma_optimality_sinkhorn_potential_appendix}, we have $\int_\XM h(x, y)\dB \beta(y) = 1$ and hence $\langle h-1, T_\sharp\alpha\otimes\beta\rangle = 0$.
	Subtracting the 1st equality from the last inequality, 
	\begin{align*}
		\OTgamma(T_\sharp \alpha, \beta) - \OTgamma(\alpha, \beta) \geq \langle f,  T_\sharp\alpha - \alpha\rangle.
	\end{align*}
	Use the change-of-variables formula of the push-forward measure to obtain
	\begin{align*}
		\frac{1}{\epsilon}\langle f,  T_\sharp\alpha - \alpha\rangle = \frac{1}{\epsilon}\int_\XM  \big((f\circ T)(x) - f(x)\big) \dB\alpha(x)
		=\int_\XM \nabla f(x+\epsilon'\phi(x)) \phi(x) \dB\alpha(x),
	\end{align*}
	where $\epsilon' \in [0, \epsilon]$ is from the mean value theorem.
	Further use the Lipschitz continuity of $\nabla f$ in Lemma \ref{lemma_lipschitz_sinkhorn_potential_gradient}, we have
	\begin{equation*}
		\lim_{\epsilon\rightarrow 0} \frac{1}{\epsilon}\langle f,  T_\sharp\alpha - \alpha\rangle = \int_\XM \nabla f(x) \phi(x) \dB\alpha(x).
	\end{equation*}
	Since $\phi\in\HM^d$, we have $\phi(x) = \langle\phi, k(x, \cdot)\rangle_{\HM^d}$ and hence
	\begin{equation*}
		\lim_{\epsilon\rightarrow0}\frac{1}{\epsilon} \big(\OTgamma(T_\sharp \alpha, \beta) - \OTgamma(\alpha, \beta)\big) \geq \langle\int \nabla f(x)k(x, \cdot)  \dB\alpha(x), \phi\rangle_{\HM^d}.
	\end{equation*}
	
	Similarly, let $f'$ and $g'$ be the Sinkhorn potentials to $\OTgamma(T_\sharp\alpha, \beta)$, using $f'\rightarrow f$ as $\epsilon\rightarrow 0$, we can have an upper bound
	\begin{equation*}
		\lim_{\epsilon\rightarrow0}\frac{1}{\epsilon}\big(\OTgamma(T_\sharp \alpha, \beta) - \OTgamma(\alpha, \beta)\big) \leq \langle\int_\XM \lim_{\epsilon\rightarrow 0}\nabla f'(x + \epsilon'\phi(x))k(x, \cdot)  \dB\alpha(x), \phi\rangle_{\HM^d}.
	\end{equation*}
	Since $\phi\in\HM^d$, we have $\|\phi\|_{2,\infty}\leq M_\HM\|\phi\|_{\HM^d} < \infty$ with $M_\HM\in\RBB_+$ being a constant.
	Using Lemma \ref{lemma_Lipschitz_continuity_of_variation}, we have that $\nabla f'$ is Lipschitz continuous with respect to the mapping
	\begin{equation*}
		\lim_{\epsilon\rightarrow 0}\|\nabla f'(x + \epsilon'\phi(x)) - \nabla f(x + \epsilon'\phi(x))\|\leq \lim_{\epsilon\rightarrow 0}\epsilon G_T\|\phi\|_{2, \infty} = 0.
	\end{equation*}
	Besides, using Lemma \ref{lemma_lipschitz_sinkhorn_potential_gradient} we have that $\nabla f$ is continuous and hence $\lim_{\epsilon\rightarrow 0}\nabla f(x + \epsilon'\phi(x)) = \nabla f(x)$.
	Consequently we have $\lim_{\epsilon\rightarrow 0}\nabla f'(x + \epsilon'\phi(x)) = \nabla f(x)$ and hence
	\begin{equation*}
	\lim_{\epsilon\rightarrow0}\frac{1}{\epsilon}\big(\OTgamma(T_\sharp \alpha, \beta) - \OTgamma(\alpha, \beta)\big) = \langle\nabla f(x)k(x, \cdot)  \dB\alpha(x), \phi\rangle_{\HM^d}.
	\end{equation*}
	From Definition \ref{definition_variation_rkhs}, we have the result of $DF_1[0]$.
	The result of $DF_2[0]$ can be obtained similarly.

\subsection{Proof of Lemma \ref{lemma_large_N}}
\label{proof_lemma_large_N}
From Proposition \ref{proposition_variation_I} and \eqref{eqn_gradient_sinkhorn_barycenter}, we recall the expression of $D\SM_{\alpha}[0]$ by
\begin{equation}
D\SM_{\alpha}[0] = \int_\XM [\frac{1}{n}\sum_{i=1}^{n}\nabla f_{\alpha, \beta_i}(x) - \nabla f_{\alpha, \alpha}(x)]  k(x, y)\dB \alpha(x),
\end{equation}
and we have $\TM[\alpha](x) = x - \eta D\SM_{\alpha}[0](x)$.
Consequently, using Corollary \ref{corollary_lipschitz_variation_mapping} we have
\begin{align*}
	\|\TM[\alpha]\|_{lip} =&\ \max_{x\neq y} \frac{\|\TM[\alpha](x) - \TM[\alpha](y)\|}{\|x-y\|}
	= \max_{x\neq y} \frac{\|x - y - \eta (D\SM_{\alpha}[0](x) - D\SM_{\alpha}[0](y))\|}{\|x-y\|}\\
	\leq&\ 1+\eta\|D\SM_{\alpha}[0]\|_{lip} \leq 1+\eta G_cG_k.
\end{align*}
The following lemma states that $\TM[\alpha]$ is Lipschitz w.r.t. $\alpha$ in terms of the bounded Lipschitz norm.
\begin{lemma} \label{lemma_appendix_i}
	For any $y\in\XM$ and any $\alpha, \alpha'\in\MM_1^+(\XM)$, we have 
	\begin{equation*}
	\|\TM[\alpha](y) - \TM[\alpha'](y)\|_{2,\infty} \leq \eta \max\{d L_f D_k + dG_c G_k, {D_k L_{bl}}\}d_{bl}(\alpha', \alpha).
	\end{equation*}
\end{lemma}
We defer the proof to Appendix \ref{proof_lemma_appendix_i}.
	Based on such lemma, for any $h$ with $\|h\|_{bl}\leq 1$, we have
	\begin{align*}
		&|\langle h, \TM[\alpha]_{\sharp}\alpha\rangle - \langle h, \TM[\alpha']_{\sharp}\alpha'\rangle| = |\langle h\circ \TM[\alpha],\alpha\rangle - \langle h\circ \TM[\alpha'],\alpha'\rangle |\\
		\leq& |\langle h\circ \TM[\alpha],\alpha\rangle - \langle h\circ \TM[\alpha],\alpha'\rangle | + |\langle h\circ \TM[\alpha],\alpha'\rangle - \langle h\circ \TM[\alpha'],\alpha'\rangle |.
	\end{align*}
	We now bound these two terms individually: For the first term,
	\begin{align*}
		|\langle h\circ \TM[\alpha],\alpha\rangle - \langle h\circ \TM[\alpha],\alpha'\rangle | \leq \|h\circ \TM[\alpha]\|_{bl} d_{bl}(\alpha, \alpha') \\
		\leq \max\{\|h\|_\infty, \|h\|_{lip}\|\TM[\alpha]\|_{lip} \}d_{bl}(\alpha, \alpha')
		&\leq (1+\eta G_cG_k)d_{bl}(\alpha, \alpha');
	\end{align*}
	And for the second term, use Lemma \ref{lemma_appendix_i} to derive
	\begin{align*}
		&\ |\langle h\circ \TM[\alpha],\alpha'\rangle - \langle h\circ \TM[\alpha'],\alpha'\rangle |\\
		\leq&\ \| h\circ \TM[\alpha] - h\circ \TM[\alpha']\|_\infty \leq \|h\|_{lip}\max_{x\in\XM}\|\TM[\alpha](x) - \TM[\alpha'](x)\| \\
		\leq&\ \eta \max\{d L_f D_k + dG_c G_k, {D_k L_{bl}}\}d_{bl}(\alpha', \alpha).
	\end{align*}
	Combining the above inequalities, we have the result
	\begin{equation*}
		d_{bl}(\TM[\alpha]_\sharp\alpha, \TM[\alpha']_\sharp\alpha')\leq(1+\eta G_cG_k+\eta \max\{d L_f D_k + dG_c G_k, {D_k L_{bl}}\})d_{bl}(\alpha', \alpha).
	\end{equation*}
\subsubsection{Proof of Lemma \ref{lemma_appendix_i}}\label{proof_lemma_appendix_i}
	Recall the definition of $\TM[\alpha](x) = x - \eta D\SM_{\alpha}[0](x)$, where the functional $\SM_{\alpha}$ is defined in \eqref{eqn_functional_per_iteration} and the Fr\'echet derivative is computed in \eqref{eqn_gradient_sinkhorn_barycenter}. For any $y\in\XM$, we have 
	\begin{align*}
	&\|\TM[\alpha](y) - \TM[\alpha'](y)\|\leq \eta\|D\SM_{\alpha}[0](y) - D\SM_{\alpha'}[0](y)\|\\
	\leq& \eta\|\int_\XM [\frac{1}{n}\sum_{i=1}^{n}\nabla f_{\alpha, \beta_i}(x) - \nabla f_{\alpha, \alpha}(x)]  k(x, y)\dB \alpha(x) - \int_\XM [\frac{1}{n}\sum_{i=1}^{n}\nabla f_{\alpha', \beta_i}(x) - \nabla f_{\alpha', \alpha'}(x)]  k(x, y)\dB \alpha'(x)\| \\
	\leq& \eta\|\int_\XM [\frac{1}{n}\sum_{i=1}^{n}\nabla f_{\alpha, \beta_i}(x) - \nabla f_{\alpha, \alpha}(x)]  k(x, y)\dB \alpha(x) - \int_\XM [\frac{1}{n}\sum_{i=1}^{n}\nabla f_{\alpha', \beta_i}(x) - \nabla f_{\alpha', \alpha'}(x)]  k(x, y)\dB \alpha(x)\| \\
	&+\eta\|\int_\XM [\frac{1}{n}\sum_{i=1}^{n}\nabla f_{\alpha', \beta_i}(x) - \nabla f_{\alpha', \alpha'}(x)]  k(x, y)\dB \alpha(x) - \int_\XM [\frac{1}{n}\sum_{i=1}^{n}\nabla f_{\alpha', \beta_i}(x) - \nabla f_{\alpha', \alpha'}(x)]  k(x, y)\dB \alpha'(x)\| \\
	=& \eta\|\int_\XM \left([\frac{1}{n}\sum_{i=1}^{n}\nabla f_{\alpha, \beta_i}(x) - \nabla f_{\alpha, \alpha}(x)]   -[\frac{1}{n}\sum_{i=1}^{n}\nabla f_{\alpha', \beta_i}(x) - \nabla f_{\alpha', \alpha'}(x)]\right)  k(x, y)\dB \alpha(x)\| \\
	&+\eta\|\int_\XM [\frac{1}{n}\sum_{i=1}^{n}\nabla f_{\alpha', \beta_i}(x) - \nabla f_{\alpha', \alpha'}(x)]  k(x, y)\dB \alpha(x) - \int_\XM [\frac{1}{n}\sum_{i=1}^{n}\nabla f_{\alpha', \beta_i}(x) - \nabla f_{\alpha', \alpha'}(x)]  k(x, y)\dB \alpha'(x)\|.
	\end{align*} 
	For the first term, use Lemma \ref{lemma_Lipschitz_continuity_of_variation_gradient} to bound
	\begin{align*}
	&\|\int_\XM \left([\frac{1}{n}\sum_{i=1}^{n}\nabla f_{\alpha, \beta_i}(x) - \nabla f_{\alpha, \alpha}(x)]   -[\frac{1}{n}\sum_{i=1}^{n}\nabla f_{\alpha', \beta_i}(x) - \nabla f_{\alpha', \alpha'}(x)]\right)  k(x, y)\dB \alpha(x)\|\\
	=& \|\int_\XM \left(\frac{1}{n}[\sum_{i=1}^{n}\nabla f_{\alpha, \beta_i}(x) - \nabla f_{\alpha', \beta_i}(x)] - \nabla f_{\alpha, \alpha}(x) + \nabla f_{\alpha', \alpha'}(x) \right)  k(x, y)\dB \alpha(x)\|\\
	\leq& {D_k L_{bl} d_{bl}(\alpha', \alpha).}
	\end{align*}
	For the second term, we bound
	\begin{align*}
	&\|\int_\XM [\frac{1}{n}\sum_{i=1}^{n}\nabla f_{\alpha', \beta_i}(x) - \nabla f_{\alpha', \alpha'}(x)]  k(x, y)\dB \alpha(x) - \int_\XM [\frac{1}{n}\sum_{i=1}^{n}\nabla f_{\alpha', \beta_i}(x) - \nabla f_{\alpha', \alpha'}(x)]  k(x, y)\dB \alpha'(x)\|\\
	\leq& \|\int_\XM [\frac{1}{n}\sum_{i=1}^{n}\nabla f_{\alpha', \beta_i}(x) - \nabla f_{\alpha', \alpha'}(x)]  k(x, y)\dB \alpha(x) - \int_\XM [\frac{1}{n}\sum_{i=1}^{n}\nabla f_{\alpha', \beta_i}(x) - \nabla f_{\alpha', \alpha'}(x)]  k(x, y)\dB \alpha'(x)\|_1 \\
	\leq& \sum_{i=1}^{d} |\int_\XM [\frac{1}{n}\sum_{i=1}^{n}\nabla f_{\alpha', \beta_i}(x) - \nabla f_{\alpha', \alpha'}(x)]_i  k(x, y)\dB \alpha(x) - \int_\XM [\frac{1}{n}\sum_{i=1}^{n}\nabla f_{\alpha', \beta_i}(x) - \nabla f_{\alpha', \alpha'}(x)]_i  k(x, y)\dB \alpha'(x)|\\
	=& \sum_{i=1}^{d} | \langle [\frac{1}{n}\sum_{i=1}^{n}\nabla f_{\alpha', \beta_i}(\cdot) - \nabla f_{\alpha', \alpha'}(\cdot)]_i  k(\cdot, y), \alpha\rangle - \langle [\frac{1}{n}\sum_{i=1}^{n}\nabla f_{\alpha', \beta_i}(\cdot) - \nabla f_{\alpha', \alpha'}(\cdot)]_i  k(\cdot, y), \alpha'\rangle| \\
	\leq& \sum_{i=1}^{d} \|[\frac{1}{n}\sum_{i=1}^{n}\nabla f_{\alpha', \beta_i}(\cdot) - \nabla f_{\alpha', \alpha'}(\cdot)]_i  k(\cdot, y)\|_{bl} d_{bl}(\alpha', \alpha).
	\end{align*}
	Therefore, we only need to bound $\sum_{i=1}^{d}\|[\frac{1}{n}\sum_{i=1}^{n}\nabla f_{\alpha', \beta_i}(x) - \nabla f_{\alpha', \alpha'}(x)]_i  k(x, y)\|_{bl}$.
	In terms of $L^\infty$ norm, we have
	\begin{equation*}
	\sum_{i=1}^{d} \|[\frac{1}{n}\sum_{i=1}^{n}\nabla f_{\alpha', \beta_i}(\cdot) - \nabla f_{\alpha', \alpha'}(\cdot)]_i  k(\cdot, y)\|_\infty \leq d D_k \|[\nabla f_{\alpha', \beta_i}]_i\|_\infty \leq d D_k G_c.
	\end{equation*}
	In terms of $\|\cdot\|_{lip}$, denote $\tilde \nabla(x) = \frac{1}{n}\sum_{i=1}^{n}\nabla f_{\alpha', \beta_i}(x) - \nabla f_{\alpha', \alpha'}(x)$. For all $x, x'\in\XM$, we have
	\begin{align*}
	&\ \frac{|[\tilde \nabla(x)]_i k(x, y) - [\tilde \nabla(x')]_i k(x', y)|}{\|x - x'\|} \\
	\leq&\ \frac{|[\tilde \nabla(x)]_i k(x, y) - [\tilde \nabla(x')]_i k(x, y)| + |[\tilde \nabla(x')]_i k(x, y) - [\tilde \nabla(x')]_i k(x', y)|}{\|x - x'\|}\\
	\leq &\ L_f D_k + G_c G_k,
	\end{align*}
	and hence $\sum_{i=1}^{d} \|[\frac{1}{n}\sum_{i=1}^{n}\nabla f_{\alpha', \beta_i}(\cdot) - \nabla f_{\alpha', \alpha'}(\cdot)]_i  k(\cdot, y)\|_{lip} \leq d L_f D_k + dG_c G_k$.
	All together, we have for any $y\in\XM$
	\begin{equation*}
	\|\TM[\alpha](y) - \TM[\alpha'](y)\| \leq \eta \max\{d L_f D_k + dG_c G_k, {D_k L_{bl}}\}d_{bl}(\alpha', \alpha).
	\end{equation*}
\subsection{Proof of Lemma \ref{lemma_sufficient_descent}} \label{proof_lemma_sufficient_descent}
We first recall a proposition from \cite{feydy2019interpolating}, which shows that the dual potentials are the variations of $\OTgamma$ w.r.t. the underlying probability measure.
\begin{definition}
	We say $h \in\CM(\XM)$ is the first-order variation of a functional $F:\MM_1^+(\XM)\rightarrow\RBB$ at $\alpha\in\MM_1^+(\XM)$ if for any displacement $\xi = \beta - \alpha$ with $\beta\in\MM_1^+(\XM)$, we have
	\begin{equation*}
	F(\alpha+t\xi) = F(\alpha) + t\langle h, \xi\rangle + o(t).
	\end{equation*}
	Further we denote $h = \nabla_\alpha F(\alpha)$.
\end{definition}
\begin{lemma} \label{lemma_variation_OTgamma}
	The first-order variation of $\OTgamma(\alpha, \beta) (\alpha\neq\beta)$ with respect to the measures $\alpha$ and $\beta$ is the corresponding Sinkhorn potential, i.e. 
	$\nabla_{(\alpha, \beta)}\OTgamma(\alpha, \beta) = (f_{\alpha, \beta}, g_{\alpha, \beta})$.
	Further, if $\alpha = \beta$, we have
	$\nabla_{\alpha} \OTgamma(\alpha, \alpha) = 2f_{\alpha, \alpha}$.
\end{lemma}
	Recall that $\alpha^{t+1} = \TM[\alpha^t]_{\sharp}\alpha^t$ where the push-forward mapping is of the form $\TM[\alpha^t](x) = x - \eta D\SM_{\alpha^t}[0](x)$ with $D\SM_{\alpha^t}[0]$ given in \eqref{eqn_gradient_sinkhorn_barycenter}.
	Using the convexity of $\SM_{\gamma}$ and Lemma \ref{lemma_variation_OTgamma}, we have
	\begin{align*}
		&\ \SM_{\gamma}(\alpha^{t+1}) - \SM_{\gamma}(\alpha^t) \\
		\leq& \langle \nabla_{\alpha} \SM_{\gamma}(\alpha)|_{\alpha = \alpha^{t+1}}, \alpha^{t+1} - \alpha^{t}\rangle && \text{\# convexity of $\SM_{\gamma}$} \\
		=& \langle \frac{1}{n}\sum_{i=1}^{n} f_{\alpha^{t+1}, \beta_i} - f_{\alpha^{t+1}, \alpha^{t+1}}, \TM[\alpha^t]_{\sharp}\alpha^t - \alpha^t\rangle && \text{\# Lemma \ref{lemma_variation_OTgamma}}\\
		=& \langle [\frac{1}{n}\sum_{i=1}^{n} f_{\alpha^{t+1}, \beta_i} - f_{\alpha^{t+1}, \alpha^{t+1}}]\circ \TM[\alpha^t] - [\frac{1}{n}\sum_{i=1}^{n} f_{\alpha^{t+1}, \beta_i} - f_{\alpha^{t+1}, \alpha^{t+1}}], \alpha^t\rangle.  && \text{\# change-of-variables}
	\end{align*}
	For succinctness, denote $\xi^{t} \defi \frac{1}{n}\sum_{i=1}^{n}f_{\alpha^{t}, \beta_i} - f_{\alpha^{t}, \alpha^{t}}$.
	Hence, we have
	\begin{align*}
		\SM_{\gamma}(\alpha^{t+1}) - \SM_{\gamma}(\alpha^t)
		\leq&\ \langle \xi^{t+1}\circ \TM[\alpha^t] - \xi^{t+1}, \alpha^t\rangle = \int \xi^{t+1}(x - \eta D\SM_{\alpha^t}[0](x)) - \xi^{t+1}(x) \dB \alpha^{t}(x) \\
		=&\ - \eta \int \langle\nabla \xi^{t+1}(x - \eta' D\SM_{\alpha^t}[0](x)),  D\SM_{\alpha^t}[0](x)\rangle  \dB \alpha^{t}(x),
	\end{align*}
	where the last equality is from the mean value theorem with $\eta' \in [0,\eta]$.
	We now bound the integral by splitting it into three terms and analyze them one by one.
	\begin{align*}
		&\int_{\XM} \langle\nabla \xi^{t+1}(x - \eta' D\SM_{\alpha^t}[0](x)),  D\SM_{\alpha^t}[0](x)\rangle  \dB \alpha^{t}(x) \\
		= &\int_{\XM} \langle\nabla \xi^{t}(x),  D\SM_{\alpha^t}[0](x)\rangle  \dB \alpha^{t}(x) && \textcircled{1}\\
		&+ \int_{\XM} \langle\nabla \xi^{t}(x - \eta' D\SM_{\alpha^t}[0](x)) - \nabla \xi^{t}(x),  D\SM_{\alpha^t}[0](x)\rangle  \dB \alpha^{t}(x) && \textcircled{2} \\
		&+ \int_{\XM} \langle\nabla \xi^{t+1}(x - \eta' D\SM_{\alpha^t}[0](x)) - \nabla \xi^{t}(x - \eta' D\SM_{\alpha^t}[0](x)),  D\SM_{\alpha^t}[0](x)\rangle  \dB \alpha^{t}(x). && \textcircled{3}
	\end{align*}
	For $\textcircled{1}$, since $D\SM_{\alpha^t}[0] \in \HM^d$, we have $D\SM_{\alpha^t}[0](x) = \langle D\SM_{\alpha^t}[0], k(x, \cdot)\rangle$ and hence
	\begin{align*}
		\int_\XM \langle\nabla \xi^{t}(x),  D\SM_{\alpha^t}[0](x)\rangle  \dB \alpha^{t}(x) =&\ \int \langle\nabla \xi^{t}(x) k(x, \cdot),  D\SM_{\alpha^t}[0]\rangle_{\HM^d} \dB \alpha^{t}(x) \\
		=&\ \|D\SM_{\alpha^t}[0]\|^2_{\HM^d} = \SB(\alpha^t, \{\beta_i\}_{i=1}^n),
	\end{align*}
	where the last equality is from the Definition \ref{definition_KSBD} and the expression of $D\SM_{\alpha}[0]$ in \eqref{eqn_gradient_sinkhorn_barycenter}.\\
	For $\textcircled{2}$, note that the summands of $\nabla \xi^t$ are of the form $\nabla f_{\alpha, \beta}$ (or $\nabla f_{\alpha, \beta}$) which is proved to be Lipschitz in Lemma \ref{lemma_lipschitz_sinkhorn_potential_gradient}.
	Consequently, we bound
	\begin{align*}
		&|\int \langle\nabla \xi^{t}(x - \eta' D\SM_{\alpha^t}[0](x)) - \nabla \xi^{t}(x),  D\SM_{\alpha^t}[0](x)\rangle  \dB \alpha^{t}(x)| \\
		\leq& \int \|\nabla \xi^{t}(x - \eta' D\SM_{\alpha^t}[0](x)) - \nabla \xi^{t}(x)\|\|D\SM_{\alpha^t}[0](x)\|  \dB \alpha^{t}(x) \\
		\leq& \int 2L_f\eta\|D\SM_{\alpha^t}[0](x)\|^2  \dB \alpha^{t}(x) && \text{\# Lemma \ref{lemma_lipschitz_sinkhorn_potential_gradient}} \\
		\leq& 2\eta L_f M_\HM^2 \|D\SM_{\alpha^t}[0]\|^2_{\HM^d} = 2\eta L_f M_\HM^2 \SB(\alpha^t, \{\beta_i\}_{i=1}^n). && \text{\# see \eqref{eqn_RKHS_norm}}
	\end{align*}
	where we use $\forall f \in \HM^d, \exists M_\HM>0$ s.t. $\|f(x)\|\leq M_\HM\|f\|_{\HM^d}, \forall x\in\XM$ in the third inequality.\\
	For $\textcircled{3}$, similar to $\textcircled{2}$, the summands of $\nabla \xi^t$ are proved to be Lipschitz in (ii) of Lemma \ref{lemma_Lipschitz_continuity_of_variation_gradient}, and hence we bound
	\begin{align*}
		&|\int \langle\nabla \xi^{t+1}(x - \eta' D\SM_{\alpha^t}[0](x)) - \nabla \xi^{t}(x - \eta' D\SM_{\alpha^t}[0](x)),  D\SM_{\alpha^t}[0](x)\rangle  \dB \alpha^{t}(x)|\\
		\leq& \int \|\nabla \xi^{t+1}(x - \eta' D\SM_{\alpha^t}[0](x)) - \nabla \xi^{t}(x - \eta' D\SM_{\alpha^t}[0](x))\|\|D\SM_{\alpha^t}[0](x)\|  \dB \alpha^{t}(x) \\
		\leq& \int \sqrt{d}\eta L_{T}\|D\SM_{\alpha^t}[0]\|_{2,\infty}\|D\SM_{\alpha^t}[0](x)\|  \dB \alpha^{t}(x) && \text{\# Lemma \ref{lemma_Lipschitz_continuity_of_variation_gradient}}\\
		\leq& 2\eta \sqrt{d}L_T M_\HM^2 \|D\SM_{\alpha^t}[0]\|^2_{\HM^d} = 2\eta\sqrt{d} L_T M_\HM^2 \SB(\alpha^t, \{\beta_i\}_{i=1}^n) && \text{\# see \eqref{eqn_RKHS_norm}}
	\end{align*}
	Combining the bounds on $\textcircled{1}, \textcircled{2}, \textcircled{3}$, we have:
	\begin{equation*}
		\SM_{\gamma}(\alpha^{t+1}) - \SM_{\gamma}(\alpha^t) \leq - \eta(1-2\eta L_f M_\HM^2-2\eta\sqrt{d} L_T M_\HM^2) \SB(\alpha^t, \{\beta_i\}_{i=1}^n),
	\end{equation*}
	which leads to the result when we set $\eta\leq \min\{\frac{1}{8L_fM_\HM^2}, \frac{1}{8\sqrt{d}L_TM_\HM^2}\}$.
\clearpage
\section{A Discussion on the Global Optimality}\label{appendix_global_optimality}
\subsection{Proof of {Theorem }\ref{thm_optimality}} \label{Proof_of_thm_optimality}
	We first show $\int_{\XM}\|\xi(x)\|^2 \dB \alpha(x)<\infty$:\\
	\begin{align*}
	\int_{\XM}\|\xi(x)\|^2 \dB \alpha(x) &= \int_{\XM} ||\frac{1}{n}\sum_{i=1}^{n} \nabla f_{\alpha, \beta_i}(x) - \nabla f_{\alpha, \alpha}(x)||^2_2 \dB \alpha(x) \\
	&= \int_{\XM} 2||\frac{1}{n}\sum_{i=1}^{n} \nabla f_{\alpha, \beta_i}(x)\|^2 + 2\|\nabla f_{\alpha, \alpha}(x)||_2^2 \dB \alpha (x) \leq 4G_f < \infty
	\end{align*}
	(i) $\SB(\alpha, \{\beta_i\}_{i=1}^n) = 0\ \&\ \supp(\alpha) =\XM$ $\Rightarrow \max_{\beta\in\MM_1^+(\XM)} \langle -\nabla_{\alpha}\SM_{\gamma}(\alpha), \beta - \alpha\rangle \leq 0$: \\
	From the integrally strictly positive definiteness of the kernel function $k(x, x')$, we have that $\int_{\XM}\|\xi(x)\|^2 \dB \alpha(x) = 0$ which implies $\nabla \xi = \frac{1}{n}\sum_{i=1}^{n} \nabla f_{\alpha, \beta_i} - \nabla f_{\alpha, \alpha}(x) = 0$ for all $x\in\supp(\alpha)$.
	Further, we have that $\xi$ is a constant function on $\XM$ by $\supp(\alpha) =\XM$.
	Since we can shift the Sinkhorn potential by a constant amount without losing its optimality, we can always ensure that $\xi$ is exactly a zero function. This implies the optimality condition of the Sinkhorn barycenter problem: $\max_{\beta\in\MM_1^+(\XM)} \langle -\nabla_{\alpha}\SM_{\gamma}(\alpha), \beta - \alpha\rangle \leq 0$.\\
	(ii) Using Theorem \ref{theorem_convergence} and (i), one directly has the result.

\subsection{Fully Supported Property of \texttt{SD} at Finite Time} \label{appendix_fully_supported}
WLOG, suppose that $c(x,y)=\infty$ if $x\notin\XM$. From the monotonicity of Lemma \ref{lemma_sufficient_descent}, the support of $\alpha^t$ will not grow beyond $\XM$. Let $p^t$ be the density function of $\alpha^t$. The density $p^{t+1}$ is given by $p^{t+1}(x) = p^t(\TM[\alpha^{t}]^{-1}(x)) \big|\det(\nabla  \TM[\alpha^{t}]^{-1}(x))\big|$, where $\TM[\alpha^{t}]$ is the mapping defined in \eqref{eqn_pushforward_mapping}. For a sufficiently small step size, the determinant is always positive. Consequently, $p^{t+1}(x) = 0$ implies $p^t(\TM[\alpha^{t}]^{-1}(x))=0$ which is impossible since $p^t$ is f.s. Therefore, $p^{t+1}$ is also a.c. and f.s.

\subsection{Review the Assumptions for Global Convergence in Previous Works} \label{appendix_previous_work_assumption}
We briefly describe the assumptions required by previous works \cite{arbel2019maximum,mroueh2019sobolev} to guarantee the global convergence to the MMD minimization problem.
We emphasize that both of these works make assumptions on the ENTIRE measure sequence.
In the following, we use $\nu_p$ to denote the target measure.

In \cite{mroueh2019sobolev}, given a measure $\nu\in\MM_1^+(\XM)$, \citet{mroueh2019sobolev} define the Kernel Derivative Gramian Embedding (KDGE) of $\nu$ by 
\begin{equation}
	D(\nu) \defi \EBB_{x\sim\nu} \left([J\Phi(x)]^\top J\Phi(x)\right),
\end{equation}
where $\Phi$ is the feature map of a given RKHS and $J\Phi$ denotes its Jacobian matrix.
Further denote the classic Kernel Mean Embedding (KME) by
\begin{equation}
	\muB(\nu) \defi \EBB_{x\sim\nu} \Phi(x).
\end{equation}
\sod requires the entire variable measure sequence $\{\nu_q\}, q\geq 0$ to satisfy for any measure $\nu_q$ such that $\delta_{p,q}\defi\muB(\nu_q) - \muB(\nu_p) \neq 0$
\begin{equation}
	D(\nu) \delta_{p,q} \neq 0.
\end{equation}

In \cite{arbel2019maximum}, \citet{arbel2019maximum} proposed two types of assumptions such that either of them leads to the global convergence of their (noisy) gradient flow algorithm.
Specifically, denote the squared weighted Sobolev semi-norm of a function $f$ in an RKHS with respect to a measure $\nu$ by $\|f\|_{\dot H(\nu)} = \int_\XM \|\nabla f(x)\|^2d \nu(x)$.
Given two probability measures on $\XM$, $\nu_p$ and $\nu_q$, define the weighted negative Sobolev distance $\|\nu_p - \nu_q\|_{\dot H(\nu)^{-1}(\nu)}$ by
\begin{equation}
	\|\nu_p - \nu_q\|_{\dot H(\nu)^{-1}(\nu)} = \sup_{f \in L_2(\nu), \|f\|_{\dot H(\nu)}\leq 1} \big|\int_{\XM}f(x)\nu_p(x) - \int_{\XM}f(x)\nu_q(x)\big|.
\end{equation}
In Proposition 7 of \cite{arbel2019maximum}, if for the entire variable measure sequence $\{\nu_q\}$ generated by their gradient flow algorithm, $\|\nu_p - \nu_q\|_{\dot H(\nu)^{-1}(\nu)}$ is always bounded, then $\nu_q$ weakly converges to $\nu_p$ under the MMD sense.\\
Further, the authors also propose another noisy gradient flow algorithm and provide its global convergence guarantee under a different assumption:
Let $f_{\nu_p, \nu_q}$ be the unnormalized witness function to $\mathrm{MMD}(\nu_p, \nu_q)$. Let $\mu$ be the standard gaussian distribution and let $\beta>0$ be a noise level.
Denote $\DM_\beta(\nu_q) \defi \EBB_{x\sim\nu_q, \mu}[\|\nabla f_{\nu_p, \nu_q}(x + \beta \mu)\|^2]$.
The noisy gradient flow algorithm globally converges if for all $n$ there exists a noise level $\beta_n$ such that
\begin{equation}
	8\lambda^2\beta_n^2 \mathrm{MMD}(\nu_p, \nu_n) \leq \DM_{\beta_n}(\nu_n),
\end{equation}
and $\sum_{i=0}^n \beta_i^2\rightarrow \infty$.
Here $\lambda$ is some problem dependent constant.

\section{Implementation}
The code to reproducing the experimental results can be found in the following link: \url{https://github.com/shenzebang/Sinkhorn_Descent}.
Our implementation is based on Pytorch and geomloss\footnote{\url{https://www.kernel-operations.io/geomloss/}}.

\end{document}